\documentclass[12pt]{article}
\usepackage{amsmath,mathrsfs}
\usepackage{graphicx,psfrag,epsf}
\usepackage{enumerate}
\usepackage{natbib}
\usepackage{epsfig}
\usepackage{amsfonts,amssymb}
\usepackage{graphicx,latexsym,longtable}
\usepackage{multirow}
\usepackage{booktabs}
\usepackage{longtable}
\usepackage{algorithmic,algorithm2e}
\usepackage[small]{caption}
\newcommand{\argmax}{\operatornamewithlimits{argmax}}

\newtheorem{theorem}{Theorem}[section]
\usepackage{mdwlist}
\newcommand{\blind}{0}

\usepackage{subfigure}
\usepackage[compact]{titlesec}

\begin{document}
\titlespacing{\section}{0pt}{8pt}{6pt}
\titlespacing{\subsection}{0pt}{8pt}{6pt}
\titlespacing{\subsubsection}{0pt}{8pt}{6pt}

\def\spacingset#1{\renewcommand{\baselinestretch}%
{#1}\small\normalsize} \spacingset{1}


\if0\blind
{
  \title{\bf Gaussian Mixture Models with Component Means Constrained in Pre-selected Subspaces}
  \author{Mu Qiao\thanks{
    Mu Qiao is Research Staff Member at IBM Almaden Research Center, San Jose, CA 95120. Email: mqiao@us.ibm.com} \hspace{0.3in}
    Jia Li\thanks{ 
    Jia Li is Professor in the Department of Statistics  and (by courtesy) Computer Science and Engineering at The Pennsylvania
State University, University Park, PA 16802. Email: jiali@stat.psu.edu
    } 
    }
  \date{}
  \maketitle
} \fi

\if1\blind
{
  \bigskip
  \bigskip
  \bigskip
  \begin{center}
    {\LARGE\bf Title}
\end{center}
  \medskip
} \fi

\bigskip
\begin{abstract}
We investigate a Gaussian mixture model (GMM) with component means constrained in a pre-selected subspace. Applications to classification and clustering are explored. An EM-type estimation algorithm is derived. We prove that the subspace containing the component means of a GMM with a common covariance matrix also contains the modes of the density and the class means. This motivates us to find a subspace by applying weighted principal component analysis to the modes of a kernel density and the class means. To circumvent the difficulty of deciding the kernel bandwidth, we acquire multiple
subspaces from the kernel densities based on a sequence of bandwidths. The GMM constrained by each subspace is estimated; and the model yielding the maximum likelihood is chosen. A dimension reduction property is proved in the sense of being informative for classification or clustering. Experiments on real and simulated data sets are conducted to examine several ways of determining the subspace and to compare with the reduced rank mixture discriminant analysis (MDA). Our new method with the simple technique of spanning the subspace only by class means often outperforms the reduced rank MDA when the subspace dimension is very low, making it particularly appealing for visualization.
\end{abstract}

\noindent%

\spacingset{1.5}


\section{Introduction} \label{sec:intro}

The Gaussian mixture model (GMM) is a popular and effective tool for
clustering and classification.  When applied to clustering, usually
each cluster is modeled by a Gaussian distribution.  Because the
cluster labels are unknown, we face the issue of estimating a GMM.  A
thorough treatment of clustering by GMM is referred
to~\citep{McLachlan:2000a}.  \cite{Hastie:1996} proposed the mixture
discriminant analysis (MDA) for classification, which assumes a GMM
for each class.  \cite{Fraley:2002} examined the roles of GMM for
clustering, classification, and density estimation.

As a probability density, GMM enjoys great flexibility comparing with
parametric distributions.  Although GMM can approximate any smooth
density by increasing the number of components $R$, the number of
parameters in the model grows quickly with $R$, especially for high
dimensional data.  The regularization of GMM has been a major research
topic on mixture models.  Early efforts focused on controlling the
complexity of the covariance matrices, partly driven by the frequent
occurrences of singular matrices in estimation~\citep{Fraley:2002}.  
More recently, it is noted that for data with very high
dimensions, a mixture model with parsimonious covariance
structures, for instance, common diagonal matrices, may still have high
complexity due to the component means alone.  Methods to regularize
the component means have been proposed from quite different
perspectives.  \cite{li06} developed the so-called two-way mixture of
Poisson distributions in which the variables are grouped and the means
of the variables in the same group within any component are assumed
identical.  The grouping of the variables reduces the number of
parameters in the component means dramatically.  
In the same spirit,
\cite{Qiao:2010} developed the two-way mixture of Gaussians.
\cite{pan07} explored the penalized likelihood method with $L_1$ norm
penalty on the component means.  The method aims at shrinking the
component means of some variables to a common value.  Variable
selection for clustering is achieved because the variables with common
means across all the components are non-informative for cluster
labels. \cite{wang08} proposed the $L_{\infty}$ norm as a penalty
instead.

In this paper, we propose another approach to regularizing the
component means in GMM, which is more along the line of reduced rank
MDA \citep{Hastie:1996} but with profound differences.  We search for
a linear subspace in which the component means reside and estimate a
GMM under such a constraint.  The constrained GMM has a dimension
reduction property.  It is proved that with the subspace restriction
on the component means and under common covariance matrices, only a
linear projection of the data with the same dimension as the subspace
matters for classification and clustering.  The method is especially
useful for visualization when we want to view data in a low
dimensional space 
which best preserves the
classification and clustering characteristics.

The idea of restricting component means to a linear subspace was first
explored in the linear discriminant analysis (LDA).  \cite{Fisher36}
proposed to find a subspace of rank $r<K$, where $K$ is the number of
classes, so that the projected class means are spread apart maximally,
The coordinates of the optimal subspace are derived by successively
maximizing the between-class variance relative to the within-class
variance, known as {\it canonical} or {\it discriminant} variables.
Although LDA does not involve the estimation of a mixture model, the
marginal distribution of the observation without the class label is a
mixture distribution.  The idea of reduced rank LDA was used by
\cite{Hastie:1996} for GMM.  It was proved in~\cite{Hastie:1996} that
reduced rank LDA can be viewed as a Gaussian maximum likelihood
solution with the restriction that the means of Gaussians lie in a
$L$-dimension subspace, i.e.,
$\mbox{rank}\{\mu_k\}_{1}^{K}=L<\mbox{max}(K-1,p)$, where $\mu_k$'s
are the means of Gaussians and $p$ is the dimension of the
data. ~\cite{Hastie:1996} extended this concept and proposed a reduced
rank version of the mixture discriminant analysis (MDA), which
performed a reduced rank weighted LDA in each iteration of the EM
algorithm.

Another related line of research is regularizing the component means of a GMM 
in a latent factor space, i.e., the use of factor analysis in GMM. It was originally 
proposed by~\cite{Ghahramani:1997} to perform concurrent clustering and dimension reduction
using mixture of factor analyzers; see also~\cite{McLachlan:2000b} and~\cite{McLachlan:2003}. 
Factor analysis was later used to regularize the component means of a GMM in each 
state of the Hidden Markov Model (HMM) for speaker verification~\citep{kenny08,povey11}. In those models,   
the total number of parameters is significantly reduced due to the regularization, which effectively prevents
over fitting. Usually the EM type of algorithm is applied to estimate the parameters and find the latent subspace. 

The role of the subspace constraining the means differs intrinsically
between our approach, the reduced rank MDA and factor analysis based mixture models, resulting in
mathematical solutions of quite different nature. Within each iteration of the EM algorithm for estimating a
GMM, the reduced rank MDA finds the subspace with a given dimension
that yields the maximum likelihood under the current partition of the
data into the mixture components. The subspace depends on the component-based clustering of data in each iteration.
Similarly, the subspaces in factor analysis based mixture models are found through the iterations of the EM algorithm, 
as part of the model estimation. However, in our method,
we treat the seek of the subspace and the estimation of the model
separately.  The subspace is fixed throughout the estimation
of the GMM.  Mathematically speaking, we try to solve the maximum
likelihood estimation of GMM under the condition that the component
means lie in a given subspace.

Our formulation of the model estimation problem allows us to exploit
multiple and better choices of density estimate when we seek the
constraining subspace.  For instance, if we want to visualize the data
in a plane while the component means are not truly constrained to a
plane, fitting a GMM with means constrained to a plane may lead to
poor density estimation.  As a result, the plane sought during the
estimation will be problematic.  It is thus sensible to find the plane
based on a density estimate without the constraint.
Afterward, we can fit a GMM under the constraint purely for the
purpose of visualization. 
Moreover, the subspace may be specified based on prior knowledge. For instance, 
in multi-dimensional data visualization,
we may already know that the component (or cluster) means of data lie in a subspace spanned by 
several dimensions of the data. Therefore, the subspace is required to be fixed. 

We propose two approaches to finding the unknown subspace. The first approach is 
the so-called {\it modal PCA (MPCA)}. We prove that the modes (local maxima) lie in the same
constrained subspace as the component means. 
We use the {\it modal EM (MEM)}
algorithm \citep{Li:2007} to find the modes. By exploiting the modes,
we are no longer restricted to the GMM as a tool for density
estimation.  Instead, we use the kernel density estimate which avoids
sensitivity to initialization.  There is an issue of choosing the
bandwidth, which is easier than usual in our framework by the
following strategy.  We take a sequence of subspaces based on density
estimates resulting from different kernel bandwidths.  We then
estimate GMMs under the constraint of each subspace and finally choose
a model yielding the maximum likelihood. Note that, each GMM is a full model for 
the original data, although the component means are constrained in a different subspace. 
We therefore can compare the estimated likelihood under each model. 
This framework in fact
extends beyond kernel density estimation.  As discussed in
\citep{Li:2007}, modes can be found using modal EM for any density in
the form of a mixture distribution.
The second approach is an extension of MPCA which exploits class means or a union set of modes 
and class means. It is easy to see that the class means also 
reside in the same constrained subspace as the component means. Comparing with 
modes, class means do not depend on the kernel bandwidth and are more robust to 
estimate. 

Experiments on the classification of several real data sets with
moderate to high dimensions show that reduced rank MDA does not always have
good performance. We do not intend to claim that our proposed method is necessarily better than 
reduced rank MDA. However, when the constraining subspace of the component means is of a very low dimension, 
we find that the proposed method with the simple technique of finding the subspace based on 
class means 
often outperforms the reduced rank MDA, which solves a discriminant subspace via 
a much more sophisticated approach. In addition, we compare our methods with standard MDA
on the data projected to the subspace containing the component means. For data with moderately high dimensions, 
our proposed method works better.  
Besides classification, our method 
easily applies to clustering.

The rest of the paper is organized as follows. 
In Section~\ref{sec:pre}, we review some background and notation. 
We present a Gaussian mixture
model with subspace constrained component means,  
the MPCA algorithm and its extension for finding the subspace 
in Section~\ref{sec:mod}. We also present several
properties of the constrained subspace, with detailed proofs in the appendix. 
In Section~\ref{sec:estimate}, we describe the estimation algorithm for the proposed model. 
Experimental results are provided in Section~\ref{sec:exp}. Finally, we conclude
and discuss future work in Section~\ref{sec:conclude}.


\section{Preliminaries and Notation} \label{sec:pre}
Let $\boldsymbol{X}=(X_1,X_2,..., X_p)^t$, where $p$ is the dimension
of the data. A sample of $\boldsymbol{X}$ is denoted by
$\boldsymbol{x}=(x_1, x_2, ..., x_p)^t$.  We present the notations for a
general Gaussian mixture model before introducing the mixture model
with component means constrained to a given subspace. Gaussian mixture
model can be applied to both classification and clustering. Let 
the class label of $\boldsymbol{X}$ be $Y \in
\mathscr{K}=\{1,2,...,K\}$. For classification purpose, the joint
distribution of $\boldsymbol{X}$ and $Y$ under a Gaussian mixture is
\begin{eqnarray}
f(\boldsymbol{X}=\boldsymbol{x},Y=k)=a_kf_k(\boldsymbol{x})=a_k\sum_{r=1}^{R_k}\pi_{kr}
\phi(\boldsymbol{x}|\boldsymbol{\mu}_{kr},\boldsymbol{\Sigma})\;,
\label{eq:genmodel}
\end{eqnarray}
where $a_k$ is the prior probability of class $k$, satisfying
$0\leq{a_k}\leq1$ and $\sum_{k=1}^Ka_k=1$, and $f_k(\boldsymbol{x})$ is the
within-class density for $\boldsymbol{X}$. $R_k$ is the number of mixture
components used to model class $k$, and the total number of mixture
components for all the classes is $R=\sum_{k=1}^KR_k$.  Let $\pi_{kr}$
be the mixing proportions for the $r$th component in class $k$,
$0\leq\pi_{kr}\leq1$, $\sum_{r=1}^{R_k}\pi_{kr}=1$.  $\phi(\cdot)$
denotes the pdf of a Gaussian distribution: $\boldsymbol{\mu}_{kr}$ is
the mean vector for component $r$ in class $k$ and
$\boldsymbol{\Sigma}$ is the common covariance matrix shared across
all the components in all the classes.  To classify a sample $\boldsymbol{X}=\boldsymbol{x}$, the Bayes classification rule is used:
$\widehat{y}={\operatorname{argmax}}_kf(Y=k|\boldsymbol{X}=\boldsymbol{x})={\operatorname{argmax}}_kf(\boldsymbol{X}=\boldsymbol{x},Y=k)$.

In the context of clustering, the Gaussian mixture
model is now simplified as
\begin{eqnarray}
f(\boldsymbol{X}=\boldsymbol{x})=\sum_{r=1}^{R}\pi_{r} \phi(\boldsymbol{x}|\boldsymbol{\mu}_{r},\boldsymbol{\Sigma})\;,
\label{eq:clsmodel}
\end{eqnarray}
where $R$ is the total number of mixture components and $\pi_{r}$ is
the mixing proportions for the $r$th component. $\boldsymbol{\mu}_{r}$
and $\boldsymbol{\Sigma}$ denote the $r$th component mean and the
common covariance matrix for all the components.  The clustering
procedure involves first fitting the above mixture model and then
computing the posterior probability of each mixture component given a
sample point. The component with the highest posterior probability is
chosen for that sample point, and all the points belonging to the same
component form one cluster.

In this work, we assume that the Gaussian component means reside in a 
given linear subspace and estimate a GMM with subspace constrained means. 
A new algorithm, namely the {\it modal PCA (MPCA)}, is proposed to find the constrained
subspace. 
The motivations of using modes to find subspace are 
outlined in Section~\ref{sec:subspace}.
Before we present MPCA, we will first introduce
the modal EM algorithm ~\citep{Li:2007} which solves the local maxima, that is, modes, of a
mixture density. 

\textbf{Modal EM:}
Given a mixture density
$f(\boldsymbol{X}=\boldsymbol{x})=\sum_{r=1}^{R}\pi_r f_r(\boldsymbol{x})$, 
as in model (\ref{eq:clsmodel}), starting from
any initial data point $\boldsymbol{x}^{(0)}$, 
the modal EM algorithm finds a mode of the density 
by alternating the following two steps
until a stopping criterion is met.  Start with $t=0$.
\begin{enumerate}
\item
Let $p_{r}=\frac{\pi_r f_r(\boldsymbol{x}^{(t)})}{f(\boldsymbol{x}^{(t)})}\, , \; r=1,...,R.$
\item
Update $\boldsymbol{x}^{(t+1)}=\argmax_{\boldsymbol{x}}\sum_{r=1}^{R}p_r \log f_r(\boldsymbol{x})$.
\end{enumerate}

The above two steps are similar to the expectation and the
maximization steps in EM~\citep{Dempster:1977}.  The first step is the
``expectation'' step where the posterior probability of each mixture
component $r$, $1\leq r\leq R$, at the current data point
$\boldsymbol{x}^{(t)}$ is computed.  The second step is the
``maximization'' step.  $\sum_{r=1}^{R}p_r \log f_r(\boldsymbol{x})$ has a
unique maximum, if the $f_r(\boldsymbol{x})$'s are normal densities.  In the
special case of a mixture of Gaussians with common covariance matrix,
that is, $f_r(\boldsymbol{x})=\phi(\boldsymbol{x}\mid \boldsymbol{\mu}_r,
\boldsymbol{\Sigma})$, we simply have
$\boldsymbol{x}^{(t+1)}=\sum_{r=1}^{R}p_r \boldsymbol{\mu}_r$.  In modal
EM, the probability density function of the data is estimated 
nonparametrically using Gaussian kernels, which are in the form of a
Gaussian mixture distribution:
\begin{eqnarray*}
f(\boldsymbol{X}=\boldsymbol{x})=\sum_{i=1}^{n} \frac{1}{n} \phi(\boldsymbol{x}\mid \boldsymbol{x}_i, \boldsymbol{\Sigma}) \, ,
\end{eqnarray*}
where the Gaussian density function is 
\[
\phi(\boldsymbol{x}\mid \boldsymbol{x}_i, \boldsymbol{\Sigma})=\frac{1}{(2\pi)^{d/2}|\boldsymbol{\Sigma}|^{1/2}}
\exp(-\frac{1}{2}(\boldsymbol{x}-\boldsymbol{x}_i)^t\boldsymbol{\Sigma}^{-1}(\boldsymbol{x}-\boldsymbol{x}_i)) \; .
\]
We use a spherical covariance matrix
$\boldsymbol{\Sigma}=diag(\sigma^2, \sigma^2, ..., \sigma^2)$.  The
standard deviation $\sigma$ is also referred to as the {\em bandwidth}
of the Gaussian kernel. When the bandwidth of Gaussian kernels increases, the
density estimate becomes smoother, and more data points tend to ascend to
the same mode. Different numbers of modes can thus be found by
gradually increasing the bandwidth of Gaussian kernels. 
The data points are grouped into one cluster if 
they climb to the same mode. We call the mode as the cluster representative. 

In ~\citep{Li:2007}, a hierarchical clustering
approach, namely, {\it Hierarchical Mode Association Clustering
(HMAC)}, is proposed based on mode association and kernel bandwidth
growth. Given a sequence of bandwidths $\sigma_1<\sigma_2<\cdots
<\sigma_\eta$, HMAC starts with every point $\boldsymbol{x_i}$ being a cluster by
itself, which corresponds to the extreme case that $\sigma_1$ approaches
0. At any bandwidth $\sigma_l(l>1)$,
the modes, that is, cluster representatives,  
obtained from the preceding bandwidth are input into the modal
EM algorithm. The modes identified then form a
new set of cluster representatives. This procedure is repeated across
all $\sigma_l$'s.  For details of HMAC, we refer interested readers to
\citep{Li:2007}. We therefore obtain modes at different levels of bandwidth by HMAC.
The clustering performed by HMAC is only for the purpose of finding modes across
different bandwidths and should not be confused with the clustering or classification based on the Gaussian mixture model we propose here. 


\section{GMM with Subspace Constrained Means}\label{sec:mod} 
The Gaussian mixture model with subspace constrained
means is presented in this section. For brevity, we focus on the constrained mixture model
in a classification set-up, since clustering can be treated as a ``one-class'' modeling and is likewise solved.

We propose to model the within-class density by a Gaussian mixture
with component means constrained to a pre-selected subspace:
\begin{eqnarray}
f_k(\boldsymbol{x})=\sum_{r=1}^{R_k}\pi_{kr} \phi(\boldsymbol{x}|\boldsymbol{\mu}_{kr},\boldsymbol{\Sigma})
\label{eq:model}
\end{eqnarray}
\begin{eqnarray}
\boldsymbol{v}_j^{t}\cdot \boldsymbol{\mu}_{k1}=\boldsymbol{v}_j^{t}\cdot \boldsymbol{\mu}_{k2}=\cdots=\boldsymbol{v}_j^{t}\cdot \boldsymbol{\mu}_{kR_k}=c_j \;,
\label{eq:c}
\end{eqnarray}
where $\boldsymbol{v}_j$'s are linearly independent vectors, $j=1,..., q$,
$q<p$, and $c_j$ is a constant, invariant to different classes. Without loss of generality, we can assume $\{\boldsymbol{v}_1, ..., \boldsymbol{v}_q\}$ span
an orthonormal basis. Augment it 
to full rank by $\{\boldsymbol{v}_{q+1}, ..., \boldsymbol{v}_p\}$.
Suppose $\boldsymbol{\nu}=\{\boldsymbol{v}_{q+1}, ..., \boldsymbol{v}_p\}$, 
$\boldsymbol{\nu}^{\perp}=\{\boldsymbol{v}_{1}, ..., \boldsymbol{v}_q\}$, and $\boldsymbol{c}=(c_1,c_2,...,c_q)^{t}$. 
Denote the projection of a vector $\boldsymbol{\mu}$ or a matrix $U$
onto an orthonormal basis $S$ by $\textbf{Proj}_{S}^{\boldsymbol{\mu}}$ or $\textbf{Proj}_{S}^U$. 
We have $\textbf{Proj}_{\boldsymbol{\nu}^{\perp}}^{\boldsymbol{\mu}_{kr}}=\boldsymbol{c}$ over all the $k$ and $r$. That is, 
the projections of all the component means $\boldsymbol{\mu}_{kr}$'s onto 
the subspace $\boldsymbol{\nu}^{\perp}$ coincide at $\boldsymbol{c}$. 
We refer to $\boldsymbol{\nu}$ as the {\it constrained subspace} where $\boldsymbol{\mu}_{kr}$'s reside (or more 
strictly, $\boldsymbol{\mu}_{kr}$'s reside in the subspace up to a translation), 
and $\boldsymbol{\nu}^{\perp}$ as the corresponding {\it null subspace}.
Suppose the dimension of the constrained subspace $\boldsymbol{\nu}$ is $d$, then $d=p-q$. 
With the constraint (\ref{eq:c}) and the assumption of a common 
covariance matrix across all the components in all the classes, 
essentially, we assume that the data within each component 
have identical distributions in the null space $\boldsymbol{\nu}^{\perp}$. 
In the following section, we will explain how to find
an appropriate constrained subspace $\boldsymbol{\nu}$. 

\subsection{Modal PCA} \label{sec:subspace}
We introduce in this section the modal PCA (MPCA) algorithm that finds a 
constrained subspace for the component means of a Gaussian mixture and the properties of the
found subspace. We prove in Appendix A the following theorem. 
\begin{theorem} 
For a Gaussian mixture model with component means constrained in a subspace
$\boldsymbol{\nu}=\{\boldsymbol{v}_{q+1}, ..., \boldsymbol{v}_p\}$, $q<p$, and a common covariance matrix across
all the components, 
the modes of the mixture density are also constrained in the same subspace $\boldsymbol{\nu}$.
\label{theorem1}
\end{theorem}
According to Theorem~\ref{theorem1}, the modes and component means of Gaussian 
mixtures reside in the same constrained subspace. 
We use the aforementioned MEM algorithm introduced in Section~\ref{sec:pre}
to find the modes of the density. 
To avoid sensitivity to initialization and the number of components, we use 
the Gaussian kernel density estimate instead of a finite mixture model for the 
density. It is well known that mixture distributions with drastically different
parameters may yield similar densities. We are thus motivated to exploit modes which 
are geometric characteristics of the densities. 

Let us denote the set of modes found by MEM under the kernel bandwidth $\sigma$ by
$\mathcal{G}=\{\mathcal{M}_{\sigma,1},
\mathcal{M}_{\sigma,2},...,\mathcal{M}_{\sigma,|\mathcal{G}|}\}$. 
A weighted principal component analysis is proposed to find the constrained subspace. 
A weight $w_{\sigma,r}$ is assigned to the $r$th mode, which is the
proportion of sample points in the entire data ascending to that mode. 
We therefore have a weighted covariance matrix of all the
modes in $\mathcal{G}$:
\[\Sigma_{\mathcal{G}}=\sum_{r=1}^{|\mathcal{G}|}w_{\sigma,r}(\mathcal{M}_{\sigma,r}-\mu_{\mathcal{G}})^T(\mathcal{M}_{\sigma,r}-\mu_{\mathcal{G}})\;,
\]
where
$\mu_{\mathcal{G}}=\sum_{r=1}^{|\mathcal{G}|}w_{\sigma,r}\mathcal{M}_{\sigma,r}$.
The principal components are then obtained by performing an eigenvalue decomposition 
on $\Sigma_{\mathcal{G}}$. Recall the dimension of the constrained subspace $\boldsymbol{\nu}$ is $d$. 
Since the leading
principal components capture the most variation in the data, we use 
the first $d$ most significant principal components to span the 
constrained subspace $\boldsymbol{\nu}$, and the remaining principal components to span the corresponding null space $\boldsymbol{\nu}^{\perp}$. 

Given a sequence of bandwidths $\sigma_1<\sigma_2<\cdots
<\sigma_\eta$, the modes 
at different levels of bandwidth can be obtained 
using the HMAC algorithm introduced in Section~\ref{sec:pre}. 
At each level, we apply the weighted
PCA to the modes, and obtain a new constrained subspace by
their first $d$ most significant principal components. In practice, if the number of modes found at 
a particular level of bandwidth is smaller than 3, we will skip the modes at that level. 
For the extreme case, when $\sigma=0$, the subspace is
actually spanned by the principal components of the original data
points.  We therefore obtain a collection of subspaces, 
$\boldsymbol{\nu}_1,...,\boldsymbol{\nu}_\eta$, resulting from a sequence of bandwidths through HMAC.

\subsection{Extension of Modal PCA} \label{sec:extension}

In this section, we propose another approach to generating the constrained subspace, which is an extension of MPCA.
Suppose the mean of class $k$ is $\mathcal{M'}_{k}$, we have
$\mathcal{M'}_{k}=\sum_{r=1}^{R_k}\pi_{kr}\boldsymbol{\mu}_{kr}$, where $\boldsymbol{\mu}_{kr}$ is the $r$th component 
in class $k$. It is easy to see that the class means lie in the same subspace as the 
Gaussian mixture component means. From Theorem~\ref{theorem1}, we know that in Gaussian 
mixtures, the modes and component means also reside in the same constrained subspace. So the class means, modes and component means all lie in the same constrained subspace. Comparing with the modes, class means are more robust to estimate.  
It is thus natural to incorporate class means to find the subspace. 
In the new approach,  
if the dimension $d$ of the constrained subspace is smaller than $K$, 
the subspace is spanned by applying weighted PCA only to class means. Otherwise, it is spanned by applying weighted PCA to a union set of modes and class means. 

Similar to modal PCA, we first assign a weight $a_{k}$ to the $k$th class mean $\mathcal{M'}_{k}$, which is the proportion of 
the number of sample points in class $k$ over the entire data, i.e., the prior probability of class $k$. Suppose the set of class means is $\mathcal{J}=
\{\mathcal{M'}_{1},\mathcal{M'}_{2}, \cdots, \mathcal{M'}_{K}\}$. 
If $d < K$, 
we have a weighted covariance matrix of all the class means: 
\[\Sigma_{\mathcal{J}}=\sum_{r=1}^{K}a_{k}(\mathcal{M'}_{r}-\mu_{\mathcal{J}})^T(\mathcal{M'}_{r}-\mu_{\mathcal{J}})\;,
\]
where
$\mu_{\mathcal{J}}=\sum_{r=1}^{K}a_{k}\mathcal{M'}_{k}$. 
An eigenvalue decomposition on $\Sigma_{\mathcal{J}}$ is then performed to obtain all the principal components. 
Similar to MPCA, the constrained subspace is spanned by the first $d$ most significant principal components.
If $d \geq K$, we will put together all the class means and modes and assign different
weights to them. Suppose $\gamma$ is a value between 0 and 100, 
we allocate a total of $\gamma$\% of weight to the class means, and the remaining 
$(100-\gamma)$\% weights allocated proportionally to the modes. That is, the weights assigned to the class mean $\mathcal{M'}_{k}$
and the mode $\mathcal{M}_{\sigma,r}$  are $\gamma a_k\%$ and $(100-\gamma)w_{\sigma,r}\%$, respectively. Then 
the weighted covariance matrix of the union set of class means and modes becomes
\[\Sigma_{\mathcal{G}\cup\mathcal{J}}=\sum_{r=1}^{K}\gamma a_k\%(\mathcal{M'}_{r}-\mu_{\mathcal{J}})^T(\mathcal{M'}_{r}-\mu_{\mathcal{J}})+
\sum_{r=1}^{|\mathcal{G}|}(100-\gamma)w_{\sigma,r}\%(\mathcal{M}_{\sigma,r}-\mu_{\mathcal{G}})^T(\mathcal{M}_{\sigma,r}-\mu_{\mathcal{G}})\;.
\]
Different weights can be allocated to the class means and the modes. For instance, if we want the class means to play a more important role in spanning subspaces, we can set $\gamma>50$.  Again, an eigenvalue decomposition is performed on $\Sigma_{\mathcal{G}\cup\mathcal{J}}$ to obtain all the principal components and 
the first $d$ most significant principal components span the constrained subspace. To differentiate this method from 
MPCA, we denote it by MPCA-MEAN.

\subsection{Dimension Reduction}\label{sec:dr}
The mixture model with component means under constraint (\ref{eq:c})
implies a dimension reduction property for the classification purpose,
formally stated below.
\begin{theorem}
  For a Gaussian mixture model with a common covariance matrix $\boldsymbol{\Sigma}$, 
  suppose all the component mean $\boldsymbol{\mu}_{kr}$'s are constrained 
  in a subspace spanned by $\boldsymbol{\nu}=\{\boldsymbol{v}_{q+1},..., \boldsymbol{v}_p\}$, $q<p$, up to a translation,  
  only a linear projection of the data $\boldsymbol{x}$ onto a subspace spanned by $\{\boldsymbol{\Sigma}^{-1}\boldsymbol{v}_j|j=q+1,...,p\}$
  (the same dimension as $\boldsymbol{\nu}$) is informative for classification.
\label{theorem2} 
\end{theorem}  
In Appendix B, we provide the detailed proof for Theorem~\ref{theorem2}.
If the common covariance matrix $\boldsymbol{\Sigma}$ is an identity matrix (or a scalar matrix), the class label $Y$ 
only depends on the projection of $\boldsymbol{x}$ onto the constrained subspace $\boldsymbol{\nu}$. However, in general, 
$\boldsymbol{\Sigma}$ is non-identity. Hence the spanning vectors, $\boldsymbol{\Sigma}^{-1}\boldsymbol{v}_j$, $j=q+1,...,p$, for the subspace informative for classification are not orthogonal in general as well. 
In Appendix B, we use the column vectors of $orth(\{\boldsymbol{\Sigma}^{-1}\boldsymbol{v}_j|j=q+1,...,p\})$
to span this subspace. To differentiate it from the constrained subspace in which the component means lie, we call it as \textit{discriminant subspace}. 
The dimension of the discriminant subspace is referred to as \textit{discriminant dimension}, which is the dimension actually needed for
classification. The discriminant subspace is of the same dimension as the constrained subspace. 
When the discriminant dimension is small, significant dimension
reduction is achieved. Our method can thus be used as a data reduction tool 
for visualization when we want to view the classification of data in a two or three dimensional space.

Although in Appendix B we prove Theorem~\ref{theorem2} in the context of classification, the proof can be easily modified to show that 
the dimension reduction property applies to clustering as well. That is, we only need the data projected onto a subspace with the same dimension as the constrained subspace $\boldsymbol{\nu}$ to compute the posterior probability of the data belonging to a component (aka cluster). 
Similarly, we name the subspace that matters for clustering as \textit{discriminant subspace} and its dimension as \textit{discriminant dimension}. 


\section{Model Estimation} \label{sec:estimate}
We will first describe in Section~\ref{sec:estimate1} the basic version of the estimation algorithm
where the constraints on the component means are characterized by (\ref{eq:c}).  A
natural extension to the constraint in (4) is to allow the constant $c_j$
to vary with the class labels, thus leading to constraint characterized in
(\ref{eq:newc}).  The corresponding algorithm is described in Section~\ref{sec:estimate2}.

\subsection{The Algorithm} \label{sec:estimate1}
Let us first summarize the work flow of our proposed method: 
\begin{enumerate}
\setlength{\itemsep}{-5pt}
\item Given a sequence of kernel bandwidths $\sigma_1<\sigma_2<\cdots
<\sigma_\eta$, apply HMAC to find the modes of the density estimation at each bandwidth $\sigma_l$.
\item Apply MPCA or MPCA-MEAN to the modes or a union set of modes and class means at each kernel bandwidth and obtain a sequence of constrained subspaces.  
\item Estimate the Gaussian mixture model with component means constrained in each subspace and select the model yielding the maximum likelihood. 
\item Perform classification on the test data or clustering on the overall data, with the selected model from Step 3. 
\end{enumerate}
Remarks:
\begin{enumerate}
\setlength{\itemsep}{-5pt}
\item In our method, the seek of subspace and the estimation of the mixture model are separate. We first search for a sequence of 
subspaces and then estimate the model constrained in each subspace separately.
\item In Step 1, the identified modes are from the density estimation of the overall data (in clustering) or the overall
training data (in classification). 
\item For MPCA-MEAN, if the dimension $d$ of the constrained subspace is smaller than $K$, 
the subspace is spanned only by class means and is therefore fixed. We do not need to choose the subspace.
\item Some prior knowledge may be exploited to yield an appropriate subspace. Then, we can estimate GMM under the constraint 
of the given subspace directly. 
\end{enumerate}

Now we will derive an EM algorithm to estimate a GMM under the constraint of 
a given subspace. The estimation method for classification is introduced first.
A common covariance matrix $\boldsymbol{\Sigma}$ is assumed across 
all the components in all the classes. 
In class $k$, the parameters
to be estimated include the class prior probability $a_{k}$, 
the mixture component prior probabilities
$\pi_{kr}$, and the Gaussian parameters $\boldsymbol{\mu}_{kr}$, $\boldsymbol{\Sigma}$, $r=1,2,...,R_k$. 
Denote the training data
by $\{(\boldsymbol{x}_i,y_i):i=1,...,n\}$. Let $n_k$ be 
the number of data points in class $k$. The total number of data points $n$ is $\sum_{k=1}^{K}n_k$.
The class prior probability $a_k$ is 
estimated by the empirical frequency $n_k/\sum_{k'=1}^Kn_{k'}$. 
The EM algorithm comprises the following two steps:
\begin{enumerate}
\setlength{\itemsep}{-10pt}
\item
\textit{Expectation-step}: Given the current parameters, 
for each class $k$, compute the component posteriori probability for each data point $\boldsymbol{x}_i$ within class $k$: 
\begin{eqnarray}
q_{i,kr}\propto \pi_{kr}\phi(\boldsymbol{x}_i|\boldsymbol{\mu}_{kr}, \boldsymbol{\Sigma})\; , \quad \mbox{subject to} \sum_{r=1}^{R_k}q_{i,kr}=1 \; .
\label{em:estep}
\end{eqnarray}  
\item 
\textit{Maximization-step}: Update $\pi_{kr}, \boldsymbol{\mu}_{kr}$,
and $\boldsymbol{\Sigma}$, which maximize the following objective function
(the $i$ subscript indicates $\boldsymbol{x}_i$ with $y_i=k$):
\begin{eqnarray}
\sum_{k=1}^K\sum_{r=1}^{R_k}\left(\sum_{i=1}^{n_k}q_{i,kr}\right)\log\pi_{kr} +
\sum_{k=1}^K\sum_{r=1}^{R_k}\sum_{i=1}^{n_k}q_{i,kr}\log \phi(\boldsymbol{x}_i|\boldsymbol{\mu}_{kr}, \boldsymbol{\Sigma})
\label{eq:obj}
\end{eqnarray}
under the constraint (\ref{eq:c}).
\end{enumerate}

In the maximization step, the optimal $\pi_{kr}$'s are not affected by
the constraint (\ref{eq:c}) and are solved separately from $\boldsymbol{\mu}_{kr}$'s and $\boldsymbol{\Sigma}$:
\begin{eqnarray}
\pi_{kr}\propto \sum_{i=1}^{n_k}q_{i,kr}\;, \quad \sum_{r=1}^{R_k} \pi_{kr}=1 \; .
\label{eq:prior}
\end{eqnarray}
Since there are no analytic solutions to $\boldsymbol{\mu}_{kr}$'s and 
$\boldsymbol{\Sigma}$ in the above constrained optimization, 
we adopt the generalized EM (GEM) algorithm.
Specifically, we use a conditional maximization approach. In every maximization step of GEM, we first fix 
$\boldsymbol{\Sigma}$, and then update the $\boldsymbol{\mu}_{kr}$'s.  Then we 
update $\boldsymbol{\Sigma}$ conditioned on the $\boldsymbol{\mu}_{kr}$'s held fixed. 
This iteration will be repeated multiple times. 

Given $\boldsymbol{\Sigma}$, solving $\boldsymbol{\mu}_{kr}$ is non-trivial. The key steps are summarized here. 
For detailed 
derivation, we refer interested readers to Appendix C. 
In constraint (\ref{eq:c}), we have $\boldsymbol{v}_j^t\cdot\boldsymbol{\mu}_{kr}=c_j$, i.e., 
identical across all the $k$ and $r$ for $j=1,...,q$. It is easy to see that $\boldsymbol{c}=(c_1,...,c_q)^t$ is equal to the projection of the mean of the overall data onto the null space $\boldsymbol{\nu}^{\perp}$. However, in practice, we do not need the value of $\boldsymbol{c}$
in the parameter estimation. Before we give the equation to solve $\boldsymbol{\mu}_{kr}$, 
let us define a few notations first.
Assume
$\boldsymbol{\Sigma}$ is non-singular and hence positive definite, we
can write
$\boldsymbol{\Sigma}=(\boldsymbol{\Sigma}^{\frac{1}{2}})^t(\boldsymbol{\Sigma}^{\frac{1}{2}})$,
where $\boldsymbol{\Sigma}^{\frac{1}{2}}$ is of full rank. 
If the eigen decomposition of
$\boldsymbol{\Sigma}$ is
$\boldsymbol{\Sigma}=V_{\boldsymbol{\Sigma}}D_{\boldsymbol{\Sigma}}
V_{\boldsymbol{\Sigma}}^{t}$, then 
$\boldsymbol{\Sigma}^{\frac{1}{2}}=D_{\boldsymbol{\Sigma}}^{\frac{1}{2}}V_{\boldsymbol{\Sigma}}^{t}$. 
Let $\boldsymbol{V}_{null}$ be a $p\times q$ orthonormal matrix $(\boldsymbol{v}_{1}, ..., \boldsymbol{v}_q)$, the column vectors of 
which span the null space $\boldsymbol{\nu}^{\perp}$. 
Suppose $\boldsymbol{B}=\boldsymbol{\Sigma}^{\frac{1}{2}}\boldsymbol{V}_{null}$. 
Perform a singular value decomposition (SVD) on $\boldsymbol{B}$, i.e., $\boldsymbol{B}=\boldsymbol{U_{B}} \boldsymbol{D_{B}} \boldsymbol{V_{B}}^{t}$, 
where $\boldsymbol{U_B}$ is a $p\times q$ matrix, the column vectors of which form an orthonormal basis for the space spanned by 
the column vectors of $\boldsymbol{B}$. Let $\hat{\boldsymbol{U}}$ be a column augmented orthonormal matrix of $\boldsymbol{U_B}$. 
Denote $\sum_{i=1}^{n_k}q_{i,kr}$ by $l_{kr}$. Let $\bar{\boldsymbol{x}}_{kr}=\sum_{i=1}^{n_k}q_{i,kr}\boldsymbol{x}_i/l_{kr}$, i.e., the weighted sample mean of the component $r$ in class $k$, and $\boldsymbol{\breve{x}}_{kr}=\hat{\boldsymbol{U}}^{t}\left(\boldsymbol{\Sigma}^{-\frac{1}{2}}\right )^{t}\cdot \bar{\boldsymbol{x}}_{kr}$. 
Define $\breve{\boldsymbol{\mu}}_{kr}^{*}$ by the following 
Eqs. (\ref{eq:tmu1}) and (\ref{eq:tmu2}): 
\begin{enumerate}
\setlength{\itemsep}{-5pt}
\item for the first $q$ coordinates, $j=1,...,q$:
\begin{eqnarray}
\breve{\mu}_{kr,j}^{*}=\frac{\sum_{k'=1}^K\sum_{r'=1}^{R_{k'}}l_{k'r'} \breve{x}_{k'r',j}}{n}
\; , \quad \mbox{identical over $r$ and $k$}\;;
\label{eq:tmu1}
\end{eqnarray}
\item for the remaining $p-q$ coordinates, $j=q+1, ..., p$:
\begin{eqnarray}
\breve{\mu}_{kr,j}^{*}=\breve{x}_{kr,j} \;.
\label{eq:tmu2}
\end{eqnarray}
\end{enumerate}
That is, the first $q$ constrained coordinates are optimized using 
component-pooled sample mean (components from all the classes) while those $p-q$ unconstrained
coordinates are optimized separately within each component using the component-wise
sample mean. Note that we abuse the term ``sample mean'' here to mean $\boldsymbol{\breve{x}}_{kr}$, instead of $\bar{\boldsymbol{x}}_{kr'}$.
In the maximization step, 
the parameter $\boldsymbol{\mu}_{kr}$ is finally solved by: 
\begin{eqnarray}
\boldsymbol{\mu}_{kr}=(\boldsymbol{\Sigma}^{\frac{1}{2}})^t
\hat{\boldsymbol{U}} \breve{\boldsymbol{\mu}}_{kr}^{*} \;. \nonumber
\label{eq:tmu3}
\end{eqnarray}
Given the $\boldsymbol{\mu}_{kr}$'s, it is easy to solve $\boldsymbol{\Sigma}$:
\begin{eqnarray}
\boldsymbol{\Sigma}=\frac{\sum_{k=1}^K\sum_{r=1}^{R_k}\sum_{i=1}^{n_k}q_{i,kr}(\boldsymbol{x}_i-\boldsymbol{\mu}_{kr})^t(\boldsymbol{x}_i-\boldsymbol{\mu}_{kr})}
{n}\;. \nonumber
\label{eq:cov}
\end{eqnarray}

To initialize the estimation algorithm, we first choose $R_k$, the
number of mixture components for each class $k$. For simplicity, an equal
number of components are assigned to each class. The constrained model
is initialized by the estimated parameters from a standard Gaussian mixture
model with the same number of components.

We have so far discussed the model estimation in a  
classification set-up. We assume a common covariance matrix and 
a common constrained subspace for all
the components in all the classes. Similar parameter estimations can also be applied to the clustering
model. Specifically, all the data are put in one ``class''. In this ``one-class'' estimation problem,
all the parameters can be estimated likewise, by omitting the ``$k$''
subscript for classes. For brevity, we skip the details here. 

\subsection{Variation of the Algorithm} \label{sec:estimate2}
We have introduced the Gaussian mixture model with component means from different classes constrained in the same subspace. 
It is natural to modify the 
previous constraint in (\ref{eq:c}) to
\begin{eqnarray}
\boldsymbol{v}_j^{t}\cdot \boldsymbol{\mu}_{k1}=\boldsymbol{v}_j^{t}\cdot \boldsymbol{\mu}_{k2}=\cdots=\boldsymbol{v}_j^{t}\cdot \boldsymbol{\mu}_{kR_k}=c_{k,j}\;, 
\label{eq:newc}
\end{eqnarray}
where $\boldsymbol{v}_j$'s are linearly independent vectors spanning an orthonormal basis, $j=1,..., q$, $q<p$, and  $c_{k,j}$
depends on class $k$. That is, 
the projections of all the component means within class $k$ onto the null space $\boldsymbol{\nu}^{\perp}$ coincide at the constant $\boldsymbol{c}_k$, where $\boldsymbol{c}_k=(c_{k,1},c_{k,2}...,c_{k,q})^t$. In the new constraint (\ref{eq:newc}), $\{\boldsymbol{v}_{1}, ..., \boldsymbol{v}_q\}$ is the same set of vectors as used in constraint (\ref{eq:c}), which spans the null space $\boldsymbol{\nu}^{\perp}$. Because $c_k$ varies with class $k$, the 
subspace in which the component means from each class reside differs from each other by a translation, that is, these subspaces are parallel. 

We train a constrained model for each class separately, and assume a common covariance matrix across all the components in all the classes.
In the new constraint (\ref{eq:newc}), $\boldsymbol{c}_k$ is actually equal to the projection of the class mean $\mathcal{M'}_{k}$ onto 
the null space $\boldsymbol{\nu}^{\perp}$. Similar to the previous estimation, in practice, we do not need the value of $\boldsymbol{c}_k$
in the parameter estimation. With the constraint (\ref{eq:newc}), essentially, the component means in each class are now constrained in a shifted subspace parallel to $\boldsymbol{\nu}$. The shifting of subspace for each class is determined by $\boldsymbol{c}_k$, or the class mean $\mathcal{M'}_{k}$.
Suppose the dimension of the constrained subspace is $d$. In general, the dimension that matters for classification in this variation of the algorithm 
is $d+K-1$, assuming that the class means already span a subspace of dimension $K-1$.

We first subtract the class specific means from the data in the training set, that is, do a class specific centering of the data. Similarly as the algorithm outlined in Section~\ref{sec:estimate}, we put all the centered data from all the classes into one training set, find all the modes under different kernel bandwidths, and then apply MPCA to generate a sequence of constrained subspaces. 
The reason that we remove the class specific means first is that they have already played a role in spanning the subspace containing all the component means. When applying MPCA, we only want to capture the dominant directions for the variation within the classes.

Comparing with the parameter estimation in Section~\ref{sec:estimate}, the only change that we need to make is that the constrained $q$ coordinates in $\breve{\boldsymbol{\mu}}_{kr}^{*}$
are now identical over $r$, but not over class $k$. For the first $q$ coordinates, $j=1,...,q$, we have: 
\begin{eqnarray}
\breve{\mu}_{kr,j}^{*}=\frac{\sum_{r'=1}^{R_k}l_{kr'} \breve{x}_{kr',j}}{n_k}
\; , \quad \mbox{identical over $r$ in class $k$}\;. \nonumber
\label{eq:tmu11}
\end{eqnarray}
That is, the first $q$ constrained coordinates are optimized using 
component-pooled sample mean in class $k$. All the other equations in the estimation remain the same.


\section{Experiments} \label{sec:exp}
In this section, we present experimental results on several real and
simulated data sets. The mixture model with subspace constrained means, reduced rank MDA, and 
standard MDA on the projection of data onto 
the constrained subspace, 
are compared for the classification of real data with moderate to high dimensions. 
We also visualize and compare the clustering results of our proposed method and the reduced rank MDA
on several simulated data sets. 

The detailed methods tested in the experiments and their name abbreviations are summarized as follows:
\begin{itemize}
\setlength{\itemsep}{2pt}
\item {\bf GMM-MPCA} \hspace{0.1cm} The mixture model with subspace constrained means, in which the subspace is obtained by MPCA. 
\vspace{-0.2cm} 
\item {\bf GMM-MPCA-MEAN} \hspace{0.1cm} The mixture model with subspace constrained means, in which the subspace is obtained by MPCA-MEAN, as introduced in Section~\ref{sec:extension}. 
\vspace{-0.2cm} 
\item {\bf GMM-MPCA-SEP} \hspace{0.1cm} The mixture model with component means constrained by separately shifted subspace for each class, 
as introduced in Section~\ref{sec:estimate1}.
\vspace{-0.2cm} 
\item {\bf MDA-RR} \hspace{0.1cm} The reduced rank mixture discriminant analysis (MDA), which is a weighted rank reduction of the full MDA.
\vspace{-0.2cm} 
\item {\bf MDA-RR-OS} \hspace{0.1cm} The reduced rank mixture discriminant analysis (MDA), which is based on optimal scoring~\citep{Hastie:1996}, 
a multiple linear regression approach.
\vspace{-0.2cm} 
\item {\bf MDA-DR-MPCA} \hspace{0.1cm} The standard MDA on the projection of data onto the same constrained subspace selected by GMM-MPCA. 
\vspace{-0.2cm} 
\item {\bf MDA-DR-MPCA-MEAN} \hspace{0.1cm} The standard MDA on the projection of data onto the same constrained subspace selected by GMM-MPCA-MEAN.
\end{itemize}
Remarks:
\begin{enumerate}
\setlength{\itemsep}{-5pt}
\item Since the most relevant work to our proposed method is reduced rank
mixture discriminant analysis (MDA), we briefly introduce MDA-RR and MDA-RR-OS in Section~\ref{sec:rrmda}.
\item In MDA-DR-MPCA or MDA-DR-MPCA-MEAN, the data are projected onto 
the constrained subspace which has yielded the largest training likelihood 
in GMM-MPCA or GMM-MPCA-MEAN. Note that 
this constrained subspace is spanned by $\boldsymbol{\nu}=\{\boldsymbol{v}_{q+1}, ..., \boldsymbol{v}_p\}$, which is found by MPCA or MPCA-MEAN, rather than the discriminant subspace informative for classification. We then apply standard MDA (assume a common covariance matrix across all the components in all the classes) to the projected training data, and classify the test data projected onto the same subspace. Note that, if we project the data onto the discriminant subspace spanned by $\{\boldsymbol{\Sigma}^{-1}\boldsymbol{v}_j|j=q+1,...,p\}$, and then apply standard MDA to classification, it is theoretically equivalent to GMM-MPCA or GMM-MPCA-MEAN (ignoring the variation caused by model estimation). The reason that we conduct these comparisons is multi-fold: first, we want to see if there is advantage of the proposed method as compared to a relative naive dimension reduction scheme; second, when the dimension of the data is high, we want to investigate if the proposed method has robust estimation of $\boldsymbol{\Sigma}$; third, 
we want to investigate the difference between the constrained subspace and the discriminant subspace. 
\end{enumerate}


\subsection{Reduced Rank Mixture Discriminant Analysis}\label{sec:rrmda}

Reduced rank MDA is a data reduction 
method which allows us to have a low dimensional view on
the classification of data in a discriminant subspace, 
by controlling the within-class spread of component means relative 
to the between class spread. We outline its estimation method
in Appendix D, which is a weighted rank reduction of the full
mixture solution proposed by \cite{Hastie:1996}. We also show how to obtain the discriminant subspace of the reduced rank method
in Appendix D. 

\citet{Hastie:1996} applied the optimal scoring approach~\citep{Breiman84} to fit
reduced rank MDA, which converted the discriminant analysis to a
nonparametric multiple linear regression problem. By expressing the
problem as a multiple regression, the fitting procedures can be
generalized using more sophisticated regression methods than linear
regression~\citep{Hastie:1996}, for instance, flexible discriminant
analysis (FDA) and penalized discriminant analysis (PDA).  The use of
optimal scoring also has some computational advantages, for instance,
using fewer observations than the weighted rank reduction. A
software package
containing a set of functions to fit MDA, FDA, and PDA by
multiple regressions is provided by \cite{Hastie:1996}. 

Although the above benefits for estimating reduced rank MDA are gained
from the optimal scoring approach, there are also some
restrictions. For instance, it can not be easily extended to fit a
mixture model for clustering since the component means and covariance
are not estimated explicitly. In addition, when the dimension of the data
is larger than the sample size, optimal scaling can not be used due to
the lack of degrees of freedom in regression.  In the following
experiment section, we will compare our proposed methods with reduced
rank MDA. Both our own implementation of reduced rank MDA based on
weighted rank reduction of the full mixture, i.e., MDA-RR, and the implementation via
optimal scoring from the software package provided by
\cite{Hastie:1996}, i.e., MDA-RR-OS, are tested.


\subsection{Classification}
Eight data sets from various sources are used for classification. We summarize the
detailed information of these data below.
\begin{itemize}
\setlength{\itemsep}{-4pt}
\item The {\bf sonar} data set consists of 208 patterns of
sonar signals.  Each pattern has 60 dimensions and the number of
classes is two. The sample sizes of the two classes are 
$(111, 97)$. 
\item The {\bf robot} data set has 5456
navigation instances, with 24 dimensions and four classes (826, 2097,
2205, 328). 
\item The {\bf waveform} data~\citep{Hastie:2001} is a
simulated three-classes data of 21 features, with a waveform
function generating both training and test sets (300, 500).  
\item The {\bf imagery} semantics data set~\citep{Qiao:2010} contains 1400 images each
represented by a 64 dimensional feature vector. These 1400 images come
from five classes with different semantics (300, 300, 300, 300,
200).
\item The {\bf parkinsons} data set is composed of 195 individual voice recordings, which are 
of 21 dimensions and divided into two classes (147, 48).
\item The {\bf satellite} data set consists of 6435 instances which are square neighborhoods of 
pixels, with 36 dimensions and six classes (1533, 703, 1358, 626, 707, 1508). 
\item The {\bf semeion} handwritten digit data have 1593 binary images from ten classes (0-9
digits) with roughly equal sample size in each class. Each image is of
$16\times16$ pixels and thus has 256 dimensions.  Four fifths of the
images are randomly selected to form a training set and the remaining
as testing. 
\item The {\bf yaleB} face image data~\citep{Georghiades01, lee05, He05} contains gray scale human
face images for 38 individuals.  Each individual has 64 images, which
are of $32\times32$ pixels, normalized to unit vectors. We randomly
select the images of five individuals, and form a data set of 250
training images and 70 test images, with equal sample size for each
individual. 
\end{itemize}

The sonar, robot, parkinsons, satellite and semeion data are from the UCI machine
learning repository. Among the above data sets,   
the semeion and yaleB data have high dimensions. The other data sets 
are of moderately high dimensions. 

For the data sets with moderately high dimensions,
five-fold cross validation is used to compute their classification
accuracy except for the waveform, whose accuracy is the average over
ten simulations, the same setting used 
in~\citep{Hastie:2001}. We assume a full common covariance matrix
across all the components in all the classes. For the semeion
and yaleB data sets, the randomly split
training and test samples are used to compute their classification
accuracy instead of cross validation due to the high computational cost. 
Since these two data sets are of high dimensions, for all the tested methods, 
we assume common diagonal covariance matrices across all the components in all the classes. 
For simplicity, the same number of mixture components is used to model each class 
for all the methods. 

In our proposed methods, the constrained subspaces are found by MPCA or MPCA-MEAN, 
introduced in Section~\ref{sec:subspace} and \ref{sec:extension}. Specifically, 
in MPCA, a sequence of
subspaces are identified from the training data by gradually increasing the
kernel bandwidth $\sigma_l$, i.e., $\sigma_1<\sigma_2<\cdots
<\sigma_\eta$, $l=1,2,...,\eta$.  In practice, we set $\eta=20$ and
choose $\sigma_l$'s equally spaced from $[0.1\hat{\sigma},
2\hat{\sigma}]$, where $\hat{\sigma}$ is the largest sample standard
deviation of all the dimensions in the data.  HMAC is used to obtain
the modes at different bandwidths. Note that in HMAC, some $\sigma_l$
may result in the same clustering as $\sigma_{l-1}$, indicating that
the bandwidth needs to be increased substantially so that the clustering 
result will be changed. 
In our experiments, only the
modes at the bandwidth resulting in different clustering from the
preceding bandwidth are employed to span the
subspace. For 
the high dimensional data, since the previous
kernel bandwidth range $[0.1\hat{\sigma}, 2\hat{\sigma}]$
does not yield a sequence of
distinguishable subspaces, we therefore increase their bandwidths.
Specifically, for the semeion and yaleB data, the kernel bandwidth $\sigma_l$ is now chosen equally spaced from $[4\hat{\sigma},
5\hat{\sigma}]$ and $[2\hat{\sigma}, 3\hat{\sigma}]$, respectively, with the interval 
being $0.1\hat{\sigma}$. In GMM-MPCA-SEP, 
since the modes are identified from a new set of class mean removed data, for both the 
semeion and yaleB data, the kernel bandwidth $\sigma_l$ is now chosen equally spaced from $[3.1\hat{\sigma},
5\hat{\sigma}]$, with the interval being $0.1\hat{\sigma}$. For the other data sets, 
$\sigma_l$ is still chosen equally spaced from $[0.1\hat{\sigma}, 2\hat{\sigma}]$.
In MPCA-MEAN, if the dimension of the constrained subspace is 
smaller than the class number $K$, the subspace is obtained by applying weighted PCA only to class means. Otherwise, 
at each bandwidth, we obtain the subspace by applying weighted PCA to a union set of class means and modes, with 60\% weight allocated 
proportionally to the means and 40\% to the modes, that is, $\gamma=60$. The subspace yielding the largest likelihood on the
training data is finally chosen as the constrained subspace.

\subsubsection{Classification Results}
We show the classification results of the tested methods in this section. 
The classification error rates on data
sets of moderately high dimensions are shown in
Tables~\ref{table11}, \ref{table12}, and \ref{table13}. We vary the discriminant dimension $d$ and also the number of mixture components
used for modeling each class. Similarly,
Table~\ref{table3} shows the classification error rates on
the semeion and yaleB data, which are of
high dimensions. For all the methods except GMM-MPCA-SEP, the dimension of the discriminant subspace equals the dimension of the constrained subspace, 
denoted by $d$. For GMM-MPCA-SEP, the dimension of the discriminant space is actually $K-1+d$. In order to compare on a common ground, 
for GMM-MPCA-SEP, we change the notation for the dimension of the constrained subspace to $d'$, and still denote the dimension of the discriminant 
subspace by $d=K-1+d'$. The minimum number of dimensions
used for classification in GMM-MPCA-SEP is therefore $K-1$.
In all these tables, if $d$ 
is set to be smaller than $K-1$, we do not have the classification results of GMM-MPCA-SEP, which are marked by ``NA''. 
In addition, in Table~\ref{table32}, the classification error rates of MDA-RR-OS on yaleB data are not reported since 
the dimension $p$ of the data is significantly larger than the sample size $n$. 
The reduced rank MDA based on optimal scoring approach cannot be applied due to the lack of degree
freedom in the regression step for the small $n$ large $p$ problem. The minimum error rate in each column is in bold font.  
From the results in these tables, we summarize our findings as follows:
\begin{itemize}
\setlength{\itemsep}{-2pt}
\item Comparing the three Gaussian mixture models with subspace constrained means, GMM-MPCA-MEAN and GMM-MPCA-SEP usually outperform GMM-MPCA, except on the waveform data. Since the class means are involved in spanning the constrained subspace in GMM-MPCA-MEAN and 
determine the shifting of the subspace for each class in GMM-MPCA-SEP, the observed advantage of 
GMM-MPCA-MEAN and GMM-MPCA-SEP indicates that class means are valuable for finding a good 
subspace. 
\item Comparing the proposed methods and the reduced rank MDA methods, when the discriminant dimension is
low, GMM-MPCA-MEAN and GMM-MPCA-SEP usually perform better than MDA-RR and MDA-RR-OS. When the discriminant dimension becomes higher, 
we do not observe a clear winner among different methods. The results are very data-dependent. Note that in GMM-MPCA-MEAN, when the discriminant dimension is smaller than $K-1$ , the subspace is obtained by applying weighted PCA only to the class means. For most data sets, when the discriminant dimension is very low, GMM-MPCA-MEAN performs best or close to best. 
\item Comparing the proposed methods and the simple methods of finding the subspace first and then fitting MDA on the data projected onto the subspace, when the data dimension is moderately high and the discriminant dimension is very low, GMM-MPCA/GMM-MPCA-MEAN usually perform better than MDA-DR-MPCA/MDA-DR-MPCA-MEAN. As the discriminant dimension increases, with certain component numbers, MDA-DR-MPCA/MDA-DR-MPCA-MEAN may have a better classification accuracy. In addition, if the data dimension is 
high, for instance, the yaleB data, MDA-DR-MPCA/MDA-DR-MPCA-MEAN may perform better even at lower discriminant dimension. As discussed in Remark 2 of this section, for MDA-DR-MPCA/MDA-DR-MPCA-MEAN and GMM-MPCA/GMM-MPCA-MEAN, we essentially do classification on the data in two different subspaces, i.e., the constrained subspace and the discriminant subspace. For GMM-MPCA/GMM-MPCA-MEAN,  under the subspace constraint, we need to estimate a common covariance matrix, which affects the discriminant subspace, as shown in Section~\ref{sec:dr}. 
Generally speaking, when the discriminant dimension becomes higher or the data dimension is high, it becomes more difficult to accurately estimate the covariance matrix. For instance, for the high dimensional data, we assume a common diagonal covariance matrix, so that the covariance estimation becomes feasible and avoids singularity issue. However, this may result in a poor discriminant subspace, which leads to worse classification accuracy. On the other hand, when the data dimension is moderately high and the discriminant dimension is very low, the estimated covariance matrix is more accurate and the discriminant subspace informative for classification
is empirically better than the constrained subspace.
\item As a final note, when the discriminant dimension is low, MDA-DR-MPCA-MEAN generally outperforms MDA-DR-MPCA. 
\end{itemize}

\begin{table}[t!]
\caption{Classification error rates (\%) for the data with moderately high dimensions (I)}
\vspace{-0.3cm}
\tiny
\begin{center}
\subtable[Robots data]{
\begin{tabular}{|cl||cccccccc|}
\hline 
\multicolumn{2}{ |c||}{Num of components} & $d=2$ & $d=5$ & $d=7$ & $d=9$ & $d=11$ & $d=13$ & $d=15$ & $d=17$ \\
\hline
\multirow{5}*{3}  
									& GMM-MPCA & 41.39 & 35.78 & 31.93 & 31.73 & 31.25 & 31.54 & 31.65 & 31.60\\ 
									& GMM-MPCA-MEAN& \textbf{30.32} & 32.06 & \textbf{30.11} & 30.52 & 31.19 & 31.40 & 31.69 & 31.29\\ 
								  & GMM-MPCA-SEP& NA & 30.86 & 30.68 & \textbf{30.42} & \textbf{29.58} & 30.28 & 30.97 & 30.68\\ 
								  & MDA-RR & 41.22 & \textbf{30.32} & 30.85 & 30.57 & 29.95 & \textbf{29.95} & \textbf{29.95} & \textbf{29.95}\\ 
								  & MDA-RR-OS & 40.16 & 32.73 & 32.44 & 30.35 & 30.66 & 30.43 & 30.26 & 31.01\\ 
								  & MDA-DR-MPCA & 44.10 & 40.30 & 35.04 & 32.72 & 33.03 & 33.56 & 33.39 & 32.84\\ 
								  & MDA-DR-MPCA-MEAN & 41.22 & 36.42 & 34.71 & 33.83 & 32.75 & 32.44 & 33.47 & 32.26\\ 
\hline
\multirow{5}*{4}  
									& GMM-MPCA & 40.74 & 31.91 & 30.77 & 30.15 & 29.40 & 29.05 & 27.91 & 28.24\\ 
									& GMM-MPCA-MEAN& \textbf{26.56} & 31.49 & \textbf{29.71} & 29.98 & 28.43 & 28.02 & 28.12 & 28.39\\
								  & GMM-MPCA-SEP& NA & \textbf{31.41} & 30.19 & 28.45 & 28.92 & 30.13 & 29.54 & 30.39\\ 
								  & MDA-RR & 40.45 & 33.63 & 30.41 & \textbf{28.28} & \textbf{27.77} & \textbf{27.09} & \textbf{27.18} & \textbf{27.18}\\ 
								  & MDA-RR-OS & 40.91& 31.87 & 31.36 & 30.24 & 27.88 & 29.01 & 28.59 & 28.61\\ 
								  & MDA-DR-MPCA & 42.26& 36.16 & 34.64 & 31.95 & 30.06 & 28.90 & 29.77 & 27.56\\
								  & MDA-DR-MPCA-MEAN & 39.41& 34.38 & 34.53 & 31.96 & 29.73 & 29.45 & 28.15 & 28.28\\ 
\hline
\multirow{5}*{5}  
									& GMM-MPCA & 37.72 & 29.67 & 29.25 & 29.31 & 27.86 & 27.91 & \textbf{26.28} & 26.21\\ 
									& GMM-MPCA-MEAN& \textbf{28.72} & 27.86 & \textbf{26.98} & 26.69 & 26.83 & \textbf{25.90} & 26.37 & \textbf{26.14}\\
								  & GMM-MPCA-SEP& NA &\textbf{26.48} & 27.05 & 27.46 & 27.22 & 26.76 & 26.74 & 27.00\\
								  & MDA-RR & 40.39 &29.01 & 26.52 & \textbf{26.08} & \textbf{26.03} & 26.61 & 26.52 & 27.09\\ 
								  & MDA-RR-OS & 39.96 &30.99 & 29.38 & 28.24 & 28.48 & 27.59 & 28.24 & 27.57\\ 
								  & MDA-DR-MPCA & 41.07 &35.69 & 32.44 & 30.86 & 29.03 & 27.99 & 28.52 & 26.34\\ 
								  & MDA-DR-MPCA-MEAN & 38.34 &33.10 & 32.18 & 30.13 & 28.56 & 26.70 & 27.05 & 26.28\\
\hline
\end{tabular}
}

\subtable[Waveform data]{
\begin{tabular}{|cl||cccccccc|}
\hline %
\multicolumn{2}{ |c||}{Num of components} & $d=2$ & $d=4$ & $d=6$ & $d=8$ & $d=10$ & $d=12$ & $d=14$ & $d=16$ \\
\hline
\multirow{5}*{3} 
								 & GMM-MPCA & 15.70 & 15.64 & 16.12 & 17.10 & 17.76 & 17.80 & 18.24 & 18.64\\
								 & GMM-MPCA-MEAN& 16.12 & 16.14 & 16.82 & 17.38 & 17.76 & 17.92 & 17.90 & 18.84\\
								 & GMM-MPCA-SEP& NA & 17.08 & 17.04 & 17.22 & 17.44 & 17.50 & 17.70 & 18.34\\  
								 & MDA-RR & 16.00 & 18.48 & 18.64 & 18.58 & 18.58 & 18.58 & 18.58 & 18.58\\ 
								 & MDA-RR-OS & 15.50 & 17.20 & 18.14 & 17.98 & 18.00 & 17.84 & 18.08 & 17.98\\
								 & MDA-DR-MPCA & \textbf{14.74} & \textbf{15.28} & 15.78 & \textbf{16.14} & \textbf{16.58} & 17.12 & 17.62 & 17.82\\
								 & MDA-DR-MPCA-MEAN & \textbf{14.74} & 15.50 & \textbf{15.76} & 16.50 & 17.00 & \textbf{16.94} & \textbf{17.26} & \textbf{17.48}\\ 
\hline
\multirow{5}*{4} 
								 & GMM-MPCA & 15.56 & 16.28 & 16.06 & 16.94 & 17.84 & \textbf{17.54} & 18.58 & 19.32\\ 
								 & GMM-MPCA-MEAN& 15.84 & 16.70 & 16.90 & 17.28 & 17.96 & 18.34 & 18.36 & 18.84\\ 
								 & GMM-MPCA-SEP& NA & 16.34 & 17.14 & 17.56 & 17.56 & 18.02 & 18.16 & \textbf{18.16}\\ 
								 & MDA-RR & 15.80 & 18.12 & 18.28 & 19.06 & 19.26 & 19.66 & 19.66 & 19.66\\ 
								 & MDA-RR-OS & 15.50 & 17.54 & 18.36 & 18.36 & 19.34 & 18.92 & 18.72 & 18.88\\ 
								 & MDA-DR-MPCA & 15.18 & 15.78 & \textbf{16.00} & \textbf{16.36} & 17.12 & 17.64 & \textbf{17.64} & 18.26\\ 
								 & MDA-DR-MPCA-MEAN & \textbf{15.12} & \textbf{15.86} & 16.16 & 16.70 & \textbf{17.00} & 17.56 & 17.66 & 18.40\\ 
\hline
\multirow{5}*{5} 
								 & GMM-MPCA & 16.44 &16.72 & 16.42 & 16.96 & 17.56 & 17.86 & 18.66 & 18.52\\ 
								 & GMM-MPCA-MEAN& 16.26 & 16.30 & 17.32 & 17.72 & 18.04 & 17.68 & 18.28 & 19.04\\ 
								 & GMM-MPCA-SEP& NA &17.24 & 16.96 & 17.32 & 17.40 & 17.66 & 17.68 & \textbf{18.30}\\ 
								 & MDA-RR & 16.76& 18.18 & 18.26 & 19.14 & 19.16 & 19.70 & 19.78 & 19.78\\ 
								 & MDA-RR-OS & 15.80& 17.78 & 18.62 & 19.02 & 19.30 & 18.92 & 18.92 & 18.40\\ 
								 & MDA-DR-MPCA & 15.34 & 15.86 & \textbf{15.98} & 16.66 & \textbf{17.16} & \textbf{16.90} & 17.90 & 18.80\\
								 & MDA-DR-MPCA-MEAN & \textbf{15.08} &\textbf{15.70} & 16.76 & \textbf{16.16} & 17.30 & 17.90 & \textbf{17.56} & 18.38\\ 
\hline	 
\end{tabular}
}
\end{center}
\label{table11}
\end{table}

\begin{table}[t!]
\caption{Classification error rates (\%) for the data with moderately high dimensions (II)}
\tiny
\vspace{-0.3cm}
\begin{center}
\subtable[Sonar data]{
\begin{tabular}{|cl||cccccccc|}
\hline %
\multicolumn{2}{ |c||}{Num of components} & $d=2$ & $d=3$ & $d=5$ & $d=7$ & $d=9$ & $d=11$ & $d=13$ & $d=15$ \\
\hline %
\multirow{7}*{3} & GMM-MPCA& 39.29 &39.78 & 24.56 & 25.48 & 24.51 & 21.61 & 21.11 & 21.13\\  
								 & GMM-MPCA-MEAN& \textbf{35.92} & 23.57 & 23.54 & 24.04 & 23.09 & 22.12 & 21.63 & 21.63\\ 
								 & GMM-MPCA-SEP& NA & 27.85 & 25.45 & 24.54 & 24.06 & 24.55 & 23.56 & 24.02\\
								 & MDA-RR& 36.48 & 28.82 & 22.08 & 22.08 & 22.08 & 22.08 & 22.08 & 22.08\\
								 & MDA-RR-OS& 45.16 &25.87 & 22.61 & 24.05 & 20.68 & 23.60 & 22.59 & 23.58\\
								 & MDA-DR-MPCA& 42.31 &38.45 & 19.71 & \textbf{18.33} & 19.77 & 22.18 & \textbf{20.71} & 17.76\\ 
								 & MDA-DR-MPCA-MEAN& 39.43 &\textbf{23.56} & \textbf{18.77} & \textbf{18.33} & \textbf{18.83} & \textbf{19.78} & 21.20 & \textbf{16.81}\\ 
								 
\hline
\multirow{7}*{4} 
								 & GMM-MPCA& 40.53 & 38.88 & \textbf{20.19} & 20.72 & 18.32 & 18.75 & 17.33 & 19.74\\ 
								 & GMM-MPCA-MEAN& \textbf{35.08} & 25.45 & 20.20 & \textbf{17.83} & \textbf{17.37} & \textbf{18.26} & \textbf{17.31} & 20.71\\ 
								 & GMM-MPCA-SEP& NA &26.51 & 22.62 & 22.16 & 20.25 & 19.75 & 20.71 & \textbf{19.25}\\
								 & MDA-RR& 46.21 & 27.91 & 23.07 & 19.27 & 19.27 & 19.27 & 19.27 & 19.27\\
								 & MDA-RR-OS& 42.80 & 26.35 & 26.44 & 19.23 & 21.62 & 22.10 & 19.25 & 22.58\\ 
								 & MDA-DR-MPCA& 37.50 & 37.42 & 22.11 & 18.33 & 18.82 & 21.21 & 21.23 & 20.26\\ 
								 & MDA-DR-MPCA-MEAN& 40.85 & \textbf{22.11} & 20.24 & 19.28 & 20.24 & 19.76 & 20.73 & 19.31\\ 

\hline
\multirow{7}*{5} 
								 & GMM-MPCA& 44.77 & 39.78 & 24.56 & 25.48 & 24.51 & 21.61 & 21.11 & 21.13\\
								 & GMM-MPCA-MEAN& \textbf{35.42} & 27.89 & \textbf{21.15} & 19.73 & \textbf{18.78} & 19.71 & \textbf{18.26} & 18.76\\
								 & GMM-MPCA-SEP& NA & 32.31 & 29.35 & 20.21 & 20.22 & 20.21 & 19.23 & 21.17\\
								 & MDA-RR& 43.70 & 27.38 & 25.91 & 22.06 & 19.67 & \textbf{19.67} & 19.67 & 19.67\\ 
								 & MDA-RR-OS& 35.55 & 29.34 & 24.86 & 22.12 & 20.68 & 22.19 & 21.18 & 20.17\\ 
								 & MDA-DR-MPCA& 36.05 & 35.07 & 21.20 & 19.29 & 20.73 & 23.09 & 20.71 & 18.18\\ 
								 & MDA-DR-MPCA-MEAN& 38.37 & \textbf{26.44} & 21.21 & \textbf{18.34} & 23.10 & 24.56 & 21.64 & \textbf{18.74}\\

\hline
\end{tabular}
}

\subtable[Imagery data]{
\begin{tabular}{|cl||cccccccc|}
\hline 
\multicolumn{2}{ |c||}{Num of components} & $d=2$ & $d=4$ & $d=6$ & $d=8$ & $d=10$ & $d=12$ & $d=14$ & $d=16$\\
\hline
\multirow{5}*{3}  
									& GMM-MPCA & 55.36 & 48.00 & 40.36 & 38.64 & 38.36 & 37.43 & 36.07 & 37.86 \\ 
									& GMM-MPCA-MEAN& \textbf{44.50} & \textbf{36.21} &36.86 & 37.07 & 36.36 & 36.79 & 36.71 & 36.14\\ 
								  & GMM-MPCA-SEP& NA & NA &\textbf{35.21} & \textbf{34.07} & 35.57 & 35.79 & 35.14 & 35.64\\ 
								  & MDA-RR & 52.57 & 43.14 &40.21 & 35.86 & 35.86 & 35.71 & 35.29 & 35.29\\ 
								  & MDA-RR-OS & 52.36 & 42.50 &38.50 & \textbf{34.07} & \textbf{35.29} & \textbf{35.50} & \textbf{34.93} & \textbf{34.79}\\ 
								  & MDA-DR-MPCA & 59.93 & 49.36 &42.21 & 41.71 & 41.00 & 39.50 & 37.00 & 38.14 \\ 
								  & MDA-DR-MPCA-MEAN & 49.36 & 44.14 &40.86 & 40.93 & 38.64 & 38.50 & 37.79 & 38.07\\ 
								  
\hline
\multirow{5}*{4}  
									& GMM-MPCA & 57.00 & 48.29 &39.79 & 38.14 & 36.57 & 36.93 & 35.64 & 36.64\\ 
									& GMM-MPCA-MEAN& \textbf{45.00} & \textbf{37.00} &39.21 & 36.57 & 35.36 & 35.43 & 35.86 & 36.14\\
								  & GMM-MPCA-SEP& NA & NA &\textbf{35.00} & \textbf{35.43} & 35.07 & 35.50 & 35.43 & 35.00\\ 
								  & MDA-RR & 52.21 & 40.64 & 38.93 & 35.79 & 37.50 & 36.50 & 35.29 & 34.86 \\ 
								  & MDA-RR-OS & 51.64 & 43.57 &37.64 & 35.50 & \textbf{34.50} & \textbf{32.36} & \textbf{34.50} & \textbf{33.64}\\ 
								  & MDA-DR-MPCA & 59.71 & 50.00 &40.14 & 40.36 & 38.29 & 37.86 & 36.29 & 37.64 \\ 
								  & MDA-DR-MPCA-MEAN & 49.71 & 42.36 & 39.71 & 39.71 & 38.64 & 37.43 & 37.21 & 37.93\\ 
\hline
\multirow{5}*{5}  
									& GMM-MPCA & 57.79& 48.50 & 40.36 & 37.57 & 37.36 & 39.07 & 36.07 & 38.29\\ 
									& GMM-MPCA-MEAN& \textbf{45.64} & \textbf{36.57} & 38.64 & 36.14 & 37.00 & 36.64 & 35.64 & 35.36\\ 
								  & GMM-MPCA-SEP& NA & NA & \textbf{35.79} & 35.14 & 34.43 & 34.36 & 35.57 & 35.43\\ 
								  & MDA-RR & 53.21& 43.36& 39.00 & 36.07 & 35.86 & 34.43 & 33.93 & 34.07\\ 
								  & MDA-RR-OS & 52.07& 42.57 & 39.71 & \textbf{34.21} & \textbf{32.64} & \textbf{34.21} & \textbf{33.50} & \textbf{32.93}\\ 
								  & MDA-DR-MPCA & 58.50 & 48.93 & 39.79 & 38.21 & 39.57 & 39.07 & 36.00 & 38.71\\ 
								  & MDA-DR-MPCA-MEAN & 50.00& 42.86 & 39.21 & 38.36 & 39.00 & 37.57 & 37.64 & 36.21\\ 
\hline
\end{tabular}
}
\end{center}
\label{table12}
\end{table}

\begin{table}[t!]
\caption{Classification error rates (\%) for the data with moderately high dimensions (III)}
\tiny
\vspace{-0.3cm}
\begin{center}
\subtable[Parkinsons data]{
\begin{tabular}{|cl||cccccccc|}
\hline %
\multicolumn{2}{ |c||}{Num of components} & $d=2$ & $d=3$ & $d=5$ & $d=7$ & $d=9$ & $d=11$ & $d=13$ & $d=15$ \\
\hline %
\multirow{7}*{3} & GMM-MPCA& 17.96 & 17.42 & 14.33 & 14.84 & 16.98 & 14.92 & 15.47 & 12.84\\ 
								 & GMM-MPCA-MEAN& 18.90 & 14.84 & \textbf{11.75} & 13.33 & 13.88 & \textbf{12.88} & 13.89 & 12.85\\ 
								 & GMM-MPCA-SEP& NA & \textbf{11.75} & 13.29 & \textbf{12.26} & 13.34 & 13.85 & \textbf{11.25} & 13.34\\ 
								 & MDA-RR& 19.42 & 15.96 & 13.88 & 13.88 & 13.88 & 13.88 & 13.88 & 13.88\\ 
								 & MDA-RR-OS& \textbf{16.88} & 16.42 & 12.31 & 13.89 & \textbf{12.31} & 13.89 & 13.37 & \textbf{12.31}\\ 
								 & MDA-DR-MPCA& 19.47 & 17.90 & 13.81 & 14.35 & 15.37 & 14.83 & 15.38 & 16.41\\ 
								 & MDA-DR-MPCA-MEAN& 19.47 & 17.90 & 13.81 & 14.33 & 15.37 & 15.35 & 15.35 & 15.35\\ 
								 
\hline
\multirow{7}*{4} 
								 & GMM-MPCA& 17.88 & 14.77 & 14.31 & 14.81 & 12.84 & 11.30 & 10.28 & 12.32\\ 
								 & GMM-MPCA-MEAN& \textbf{14.81} & 14.31 & 13.83 & 12.81 & \textbf{9.25} & \textbf{9.28} & \textbf{8.74} & 10.76\\
								 & GMM-MPCA-SEP& NA & \textbf{11.28} & 11.33 & 12.29 & 11.80 & 10.29 & 9.76 & \textbf{9.79}\\
								 & MDA-RR& 16.85 & 12.81 & 11.79 & \textbf{10.29} & 9.79 & 9.79 & 9.79 & \textbf{9.79}\\ 
								 & MDA-RR-OS& 18.41 & 15.38 & \textbf{10.74} & 10.79 & 11.84 & 11.83 & 12.85 & 10.32\\ 
								 & MDA-DR-MPCA& 19.47 & 18.47 & 12.29 & 12.35 & 10.77 & 11.23 & 10.72 & 12.30\\ 
								 & MDA-DR-MPCA-MEAN& 19.47 & 17.43 & 12.29 & 11.81 & 9.72 & 10.24 & 10.72 & 11.76\\ 

\hline
\multirow{7}*{5} 
								 & GMM-MPCA& 19.39 & 18.39 & 14.84 & 17.37 & 15.31 & 12.81 & 11.25 & 12.79\\ 
								 & GMM-MPCA-MEAN& \textbf{18.39} & 16.34 & 13.26 & 14.83 & 11.78 & 11.25 & 10.25 & 11.29\\ 
								 & GMM-MPCA-SEP& NA & \textbf{14.81} & \textbf{10.75} & 12.25 & 11.75 & \textbf{10.75} & 10.25 & 10.78\\
								 & MDA-RR& 19.94 & 16.30 & 12.27 & 13.83 & 11.28 & 10.78 & 11.28 & 10.79\\ 
								 & MDA-RR-OS& 18.96 & 16.43 & 14.30 & \textbf{11.22} & 10.25 & 12.33 & \textbf{9.71} & \textbf{9.74}\\
								 & MDA-DR-MPCA& 18.96 & 18.47 & 13.80 & 11.81 & 11.28 & 12.77 & 10.70 & 10.76\\
								 & MDA-DR-MPCA-MEAN& 18.96 & 19.52 & 12.26 & 11.31 & \textbf{9.75} & 12.27 & 10.20 & \textbf{9.74}\\ 

\hline
\end{tabular}
}
\subtable[Satellite data]{
\begin{tabular}{|cl||cccccccc|}
\hline 
\multicolumn{2}{ |c||}{Num of components} & $d=2$ & $d=4$ & $d=7$ & $d=9$ & $d=11$ & $d=13$ & $d=15$ & $d=17$ \\
\hline
\multirow{5}*{3}  
									& GMM-MPCA & \textbf{16.74} & 15.01 & 14.16 & 14.67 & 14.06 & 13.95 & 13.97 & 13.63\\ 
									& GMM-MPCA-MEAN& 16.94 & 14.10 & 13.53 & 13.77 & 13.95 & 13.78 & 13.66 & 13.68\\ 
								  & GMM-MPCA-SEP& NA & NA & 15.48 & 13.58 & 13.80 & 13.58 & 13.71 & 13.67\\
								  & MDA-RR & 35.18 & 14.41 & 12.84 & \textbf{12.96} & 13.46 & 13.60 & 13.66 & 13.53\\
								  & MDA-RR-OS & 34.90 & \textbf{13.95} & \textbf{13.01} & 13.09 & \textbf{12.82} & \textbf{13.04} & \textbf{13.35} & 13.29\\ 
								  & MDA-DR-MPCA & 17.20 & 14.83 & 13.61 & 13.91 & 13.80 & 13.35 & 13.60 & 13.69\\ 
								  & MDA-DR-MPCA-MEAN & 17.09 & 14.42 & 13.58 & 14.12 & 14.06 & 13.41 & 13.38 & \textbf{13.13}\\ 
\hline
\multirow{5}*{4}  
									& GMM-MPCA & \textbf{17.02} & 14.13 & 13.61 & 13.80 & 13.58 & 12.90 & 12.93 & 12.88\\ 
									& GMM-MPCA-MEAN& 17.31 & 13.41 & 13.50 & 13.53 & 13.24 & 12.94 & 12.91 & 12.87\\ 
								  & GMM-MPCA-SEP& NA & NA & 15.40 & 12.93 & 13.08 & 13.35 & 13.38 & 13.54 \\ 
								  & MDA-RR & 35.06 & \textbf{13.35} & 12.60 & 12.77 & 12.74 & 12.73 & 12.63 & 13.05\\ 
								  & MDA-RR-OS & 34.28 & 13.49 & \textbf{11.95} & \textbf{12.17} & \textbf{11.90} & 12.49 & \textbf{11.97} & \textbf{12.14}\\ 
								  & MDA-DR-MPCA & 17.54 & 14.14 & 13.21 & 13.52 & 13.05 & 12.74 & 12.46 & 12.45\\ 
								  & MDA-DR-MPCA-MEAN & 17.37 & 13.36 & 13.53 & 13.57 & 13.07 & \textbf{12.43} & 12.45 & 12.82\\ 
\hline
\multirow{5}*{5}  
									& GMM-MPCA & \textbf{16.25} & 13.66 & 12.90 & 13.29 & 12.79 & 12.26 & 11.92 & 12.24\\ 
									& GMM-MPCA-MEAN& 16.77 & 12.93 & 12.85 & 12.96 & 12.40 & 12.18 & 11.89 & 12.24\\ 
								  & GMM-MPCA-SEP& NA & NA & 15.48 & 13.21 & 12.56 & 12.70 & 12.59 & 12.49\\ 
								  & MDA-RR & 27.43 & 13.27 & 12.85 & 12.34 & 12.15 & 12.18 & 12.29 & 12.28\\ 
								  & MDA-RR-OS & 30.16 & 13.30 & \textbf{12.31} & \textbf{12.23} & \textbf{11.73} & \textbf{11.89} & 11.98 & \textbf{11.97}\\ 
								  & MDA-DR-MPCA & 16.61 & 13.80 & 12.70 & 12.82 & 12.74 & 12.09 & \textbf{11.79} & 12.42\\ 
								  & MDA-DR-MPCA-MEAN & 16.58 & \textbf{12.82} & 12.99 & 12.93 & 12.66 & 11.92 & 11.92 & 12.31\\ 
\hline
\end{tabular}
}

\end{center}
\label{table13}
\end{table}


\subsection{Sensitivity of Subspace to Bandwidths}
Different kernel bandwidths may result in different sets of modes by HMAC, which again may yield different constrained subspaces. 
We investigate in this section the sensitivity of constrained subspaces to kernel bandwidths. 

Assume two subspaces 
$\boldsymbol{\nu}_1$ and $\boldsymbol{\nu}_2$ are spanned by two sets of 
orthonormal basis vectors $\{\boldsymbol{v}_{1}^{(1)}, ..., \boldsymbol{v}_d^{(1)}\}$
and $\{\boldsymbol{v}_{1}^{(2)}, ..., \boldsymbol{v}_d^{(2)}\}$, where $d$
is the dimension. 
To measure the closeness between two subspaces, 
we project the basis of one subspace onto the other. Specifically, the 
closeness between $\boldsymbol{\nu}_1$ and $\boldsymbol{\nu}_2$ is defined as 
$closeness(\boldsymbol{\nu}_1, \boldsymbol{\nu}_2)=\sum_{i=1}^{d}\sum_{j=1}^{d}(\boldsymbol{v}_{i}^{(1)t}\cdot\boldsymbol{v}_{j}^{(2)})^2$. 
If $\boldsymbol{\nu}_1$ and $\boldsymbol{\nu}_2$ span the same subspace, $\sum_{j=1}^{d}(\boldsymbol{v}_{i}^{(1)t}\cdot\boldsymbol{v}_{j}^{(2)})^2=1$,
for $i=1,2,...,d$.  If they are orthogonal to
each other, $\sum_{j=1}^{d}(\boldsymbol{v}_{i}^{(1)t}\cdot\boldsymbol{v}_{j}^{(2)})^2=0$,
for $i=1,2,...,d$. Therefore, the range of 
$closeness(\boldsymbol{\nu}_1, \boldsymbol{\nu}_2)$ is $(0,d)$. The higher the value, the closer the two subspaces are. 

\begin{table}[t!]
\caption{Classification error rates (\%) for the data with high dimensions}
\vspace{-0.3cm}
\tiny
\begin{center}
\subtable[Semeion data]{
\begin{tabular}{|cl||cccccccc|}
\hline 
\multicolumn{2}{ |c||}{Num of components} & $d=2$ & $d=4$ & $d=8$ & $d=11$ & $d=13$ & $d=15$ & $d=17$ & $d=19$\\
\hline
\multirow{5}*{3} 
								 & GMM-MPCA & 53.56 & 29.72 & 18.27 & 19.81 & 18.89 & 19.20 & 18.89 & 18.27 \\ 
								 & GMM-MPCA-MEAN& 49.54 & 29.10 & 13.31 & 14.86 & 16.72 & 14.86 & 16.72 & 16.10 \\ 
								 & GMM-MPCA-SEP& NA & NA & NA & 13.31 & 12.07 & 13.00 & 16.10 & 14.86 \\ 
								 & MDA-RR & \textbf{45.51} & 26.93 & 15.79 & 14.86 & 13.93 & 15.79 & 14.24 & 13.62 \\
								 & MDA-RR-OS & 48.36 &\textbf{24.92} & 13.93 & \textbf{12.41} & \textbf{10.60} & \textbf{11.10} & \textbf{10.59} & \textbf{11.41}\\ 
								 & MDA-DR-MPCA & 49.54 & 27.86 & 19.20 & 17.03 & 17.03 & 15.79 & 16.72 & 16.41\\
								 & MDA-DR-MPCA-MEAN & 48.30 & 26.01 & \textbf{12.07} & 13.62 & 13.31 & 12.07 & 14.24 & 14.55\\

\hline
\multirow{5}*{4} 
								 & GMM-MPCA & 53.56 & 26.32 & 17.03 & 16.10 & 16.10 & 16.10 & 16.10 & 15.17\\ 
								 & GMM-MPCA-MEAN& 51.39 & 25.70 & \textbf{11.46} & 11.76 & 12.69 & 13.00 & 14.24 & 15.17\\ 
								 & GMM-MPCA-SEP& NA & NA & NA& 13.31 & 12.07 & 12.07 & 13.62 & 11.76\\ 
								 & MDA-RR & 49.23 & 26.01 & 13.93 & 13.62 & 13.00 & 12.07 & 13.62 & 12.07\\ 
								 & MDA-RR-OS & 48.70 & \textbf{24.83} & 14.21 & \textbf{11.60} & \textbf{10.59} & \textbf{11.16} & \textbf{9.97} & \textbf{11.09}\\ 
								 & MDA-DR-MPCA & 46.75 & 26.32 & 17.34 & 16.41 & 16.41 & 15.17 & 16.10 & 15.17\\ 
								 & MDA-DR-MPCA-MEAN & \textbf{44.58} & 26.32 & 13.00 & 11.76 & 15.17 & 13.31 & 13.00 & 13.00 \\ 
								 
\hline
\multirow{5}*{5} 
								 & GMM-MPCA & 51.70 & \textbf{24.46} & 15.79 & 13.62 & 15.17 & 15.17 & 13.93 & 13.00\\
								 & GMM-MPCA-MEAN& \textbf{43.03} & 26.63 & 11.15& 11.46 & 12.38 & 12.07 & 13.31 & 13.00\\ 
								 & GMM-MPCA-SEP& NA & NA& NA& 13.00 & 11.46 & 13.00 & 12.38 & 13.00\\
								 & MDA-RR & 48.92 & 25.39 & 13.00 & 12.69 & 11.46 & \textbf{10.53} & 12.07 & 12.69\\ 
								 & MDA-RR-OS & 49.16 & 26.53& 14.21 & 11.10 & \textbf{10.60} & \textbf{10.53} & 9.84 & 9.96\\ 
								 & MDA-DR-MPCA & 48.61 & 27.24 & 18.58 & 13.93 & 14.24 & 13.62 & 13.93 & 10.84\\ 
								 & MDA-DR-MPCA-MEAN & 46.13 & 25.08 & \textbf{10.22} & \textbf{10.84} & 10.84 & 10.53 & \textbf{8.98} & \textbf{9.29}\\ 
								 
\hline
\end{tabular}
\label{table31}
}

\subtable[YaleB data]{
\begin{tabular}{|cl||cccccccc|}
\hline 
\multicolumn{2}{ |c||}{Num of components} & $d=2$ & $d=4$ & $d=6$ & $d=8$ & $d=10$ & $d=12$ & $d=14$ & $d=16$ \\
\hline
\multirow{4}*{3} 
								 & GMM-MPCA & 84.29 & 64.29 & 64.29 & 55.71 & 45.71 & 38.57 & 40.00 & 34.29 \\
								 & GMM-MPCA-MEAN& 31.43 & \textbf{17.14} & 52.86 & 51.43 & 38.57 & 30.00 & 28.57 & 27.14 \\
								 & GMM-MPCA-SEP& NA & NA & \textbf{27.14} & 20.00 & \textbf{21.43} & 20.00 & 20.00 & 20.00\\
								 & MDA-RR & 87.14 & 42.86 & \textbf{27.14} & \textbf{17.14} & 28.57 & \textbf{8.57} & \textbf{11.43} & \textbf{11.43}\\ 
								 & MDA-DR-MPCA & 82.86 & 58.57 & 50.00 & 44.29 & 37.14 & 42.86 & 25.71 & 32.86 \\
								 & MDA-DR-MPCA-MEAN & \textbf{30.00} & \textbf{17.14} & 60.00 & 37.14 & 40.00 & 21.43 & 17.14 & 14.29\\ 
\hline
\multirow{4}*{4} 
								 & GMM-MPCA & 84.29 & 67.14 & 68.57 & 55.71 & 44.29 & 44.29 & 40.00 & 37.14 \\
								 & GMM-MPCA-MEAN& 34.29& 22.86 &64.29 & 50.00 & 35.71 & 30.00 & 35.71 & 30.00 \\ 
								 & GMM-MPCA-SEP& NA & NA &\textbf{31.43} & 25.71 & 28.57 & 27.14 & 24.29 & 25.71\\ 
								 & MDA-RR & 85.71& 60.00 & 41.43 & \textbf{24.29} & \textbf{14.29} & \textbf{10.00} & 12.86 & \textbf{11.43}\\ 
								 & MDA-DR-MPCA & 90.00 & 55.71 & 50.00 & 42.86 & 37.14 & 41.43 & 28.57 & 27.14\\ 
								 & MDA-DR-MPCA-MEAN & \textbf{25.71} & \textbf{21.43} & 60.00 & 35.71 & 32.86 & 11.43 & \textbf{11.43} & 12.86\\ 
\hline
\multirow{4}*{5} 
								 & GMM-MPCA & 85.71 & 65.71 & 65.71 & 55.71 & 50.00 & 45.71 & 42.86 & 40.00 \\
								 & GMM-MPCA-MEAN& 37.14 & \textbf{14.29} & 60.00 & 51.43 & 47.14 & 42.86 & 41.43 & 38.57\\
								 & GMM-MPCA-SEP& NA & NA& \textbf{31.43} & 35.71 & \textbf{30.00} & 32.86 & 28.57 & 35.71\\
								 & MDA-RR & 85.71 & 61.43 & 42.86 & 42.86 & 32.86 & 30.00 & 22.86 & 24.29 \\
								 & MDA-DR-MPCA & 87.14 & 67.14 & 52.86 & 38.57 & 34.29 & 34.29 & \textbf{17.14} & 22.86\\ 
								 & MDA-DR-MPCA-MEAN & \textbf{27.14} & 18.57 & 50.00 & \textbf{34.29} & \textbf{27.14} & \textbf{21.43} & 20.00 & \textbf{7.14}\\ 
\hline
\end{tabular}
\label{table32}
}
\end{center}
\label{table3}
\end{table}

In our proposed methods, a collection of constrained subspaces are obtained through MPCA or MPCA-MEAN
at different kernel bandwidth $\sigma_l$'s, $l=1,2,...,\eta$, and $\sigma_1<\sigma_2<\cdots<\sigma_\eta$. 
To measure the sensitivity of subspaces to different bandwidths, we compute the mean closeness between the subspace 
found at $\sigma_l$ and all the other subspaces at preceding bandwidths $\sigma_{l'},l'=1,2,...,l-1$. A large mean closeness 
indicates that the current subspace is close to preceding subspaces. 
Table~\ref{table:sensitivity1} lists
the mean closeness of subspaces by MPCA and MPCA-MEAN 
at different bandwidth levels for the sonar and imagery data (the training set from one fold in the previous five-fold cross validation setup). 
The first values in the parentheses are by MPCA while the second values are by MPCA-MEAN. 
We vary the dimension of the constrained subspace. 
The number of modes identified at each level is also shown in the tables. 
As Table~\ref{table:sensitivity1} shows, for both methods, 
the subspaces found at the first few levels are close to each other, indicated by their large mean closeness values, which are close 
to $d$, the dimension of the subspace. As the bandwidth $\sigma_l$ increases, the mean closeness starts to decline, which indicates that the 
corresponding subspace changes. When $\sigma_l$ is small, the number of modes identified by HMAC is large. The modes and their
associated weights do not change much. As a result, the generated subspaces at these bandwidths are relatively stable. As $\sigma_l$ increases, the kernel density estimate becomes smoother, and more data points tend to ascend to the same mode. We thus have a smaller number of modes with 
changing weights, which may yield a substantially different subspace. Additionally, the subspace by MPCA-MEAN is spanned by applying weighted PCA to a union set of modes and class means. In our experiment, we have allocated a larger weight proportionally to class means (in total, 60\%) and the class means remain unchanged in the union set at each kernel bandwidth. Therefore, the differences between subspaces by MPCA-MEAN are smaller than that by MPCA, indicated by larger closeness values. 


\subsection{Model Selection}
In our proposed method, the following model selection strategy is adopted. We  
take a sequence of subspaces resulting from different kernel bandwidths, and then 
estimate a mixture model constrained by each subspace and finally choose a model yielding the maximum likelihood.
In this section, we examine our model selection criteria, and the 
relationships among test classification error rates, training likelihoods and kernel bandwidths. 

\begin{figure}[t]
\centering
\begin{tabular}{cc}
\epsfig{file=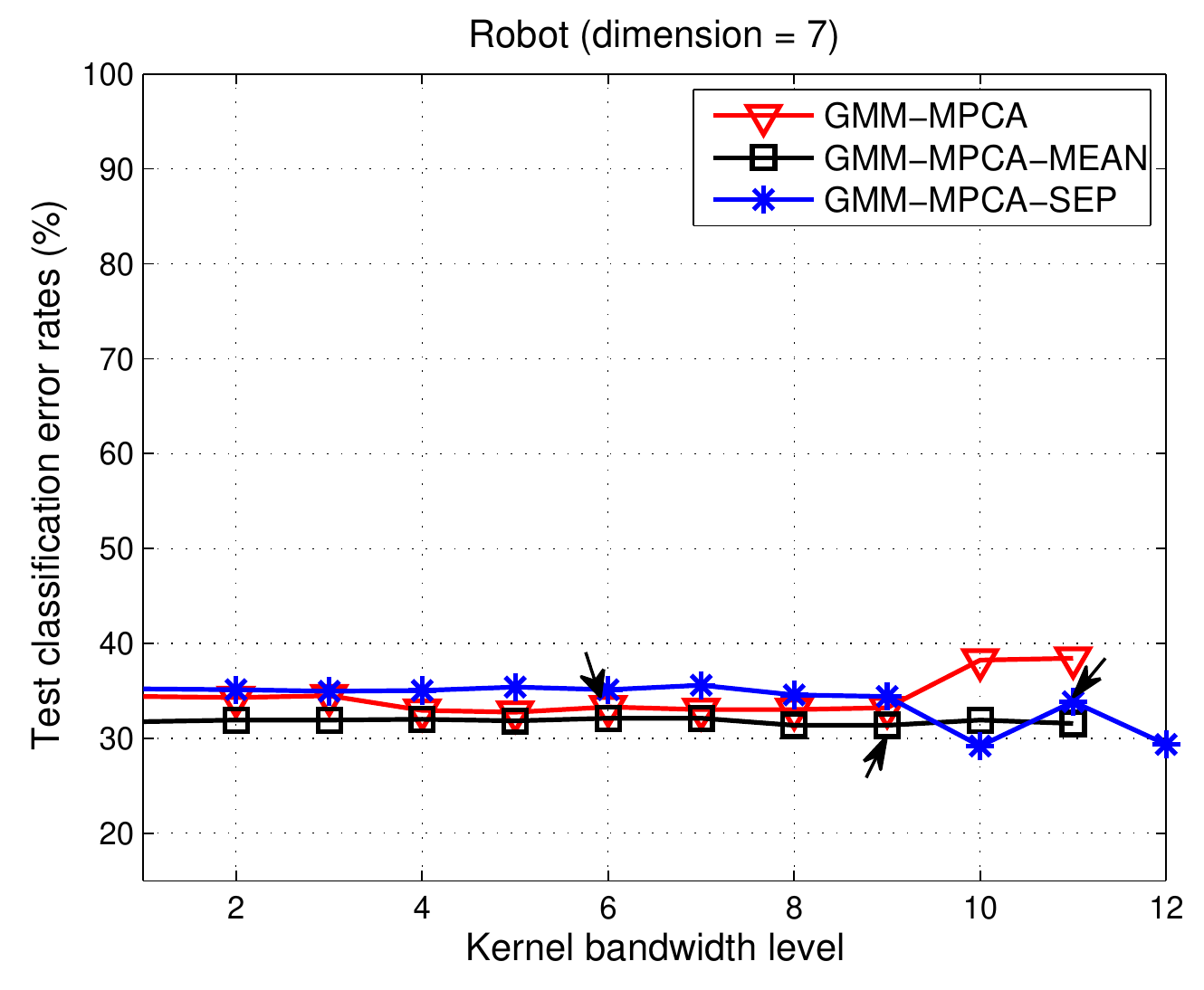,width=2.4in, origin=br,angle=0}
\epsfig{file=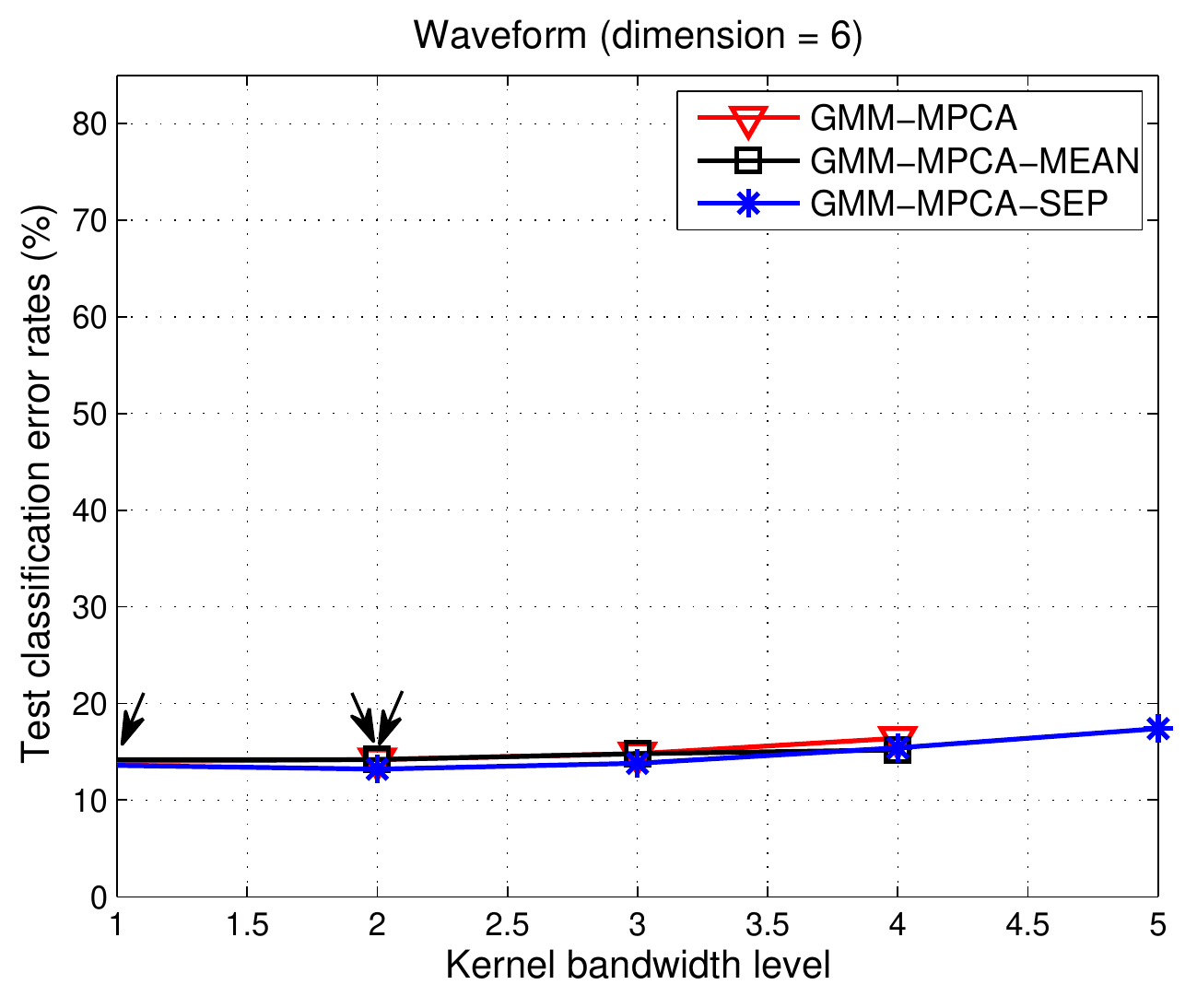,width=2.37in, origin=br,angle=0} \\
\epsfig{file=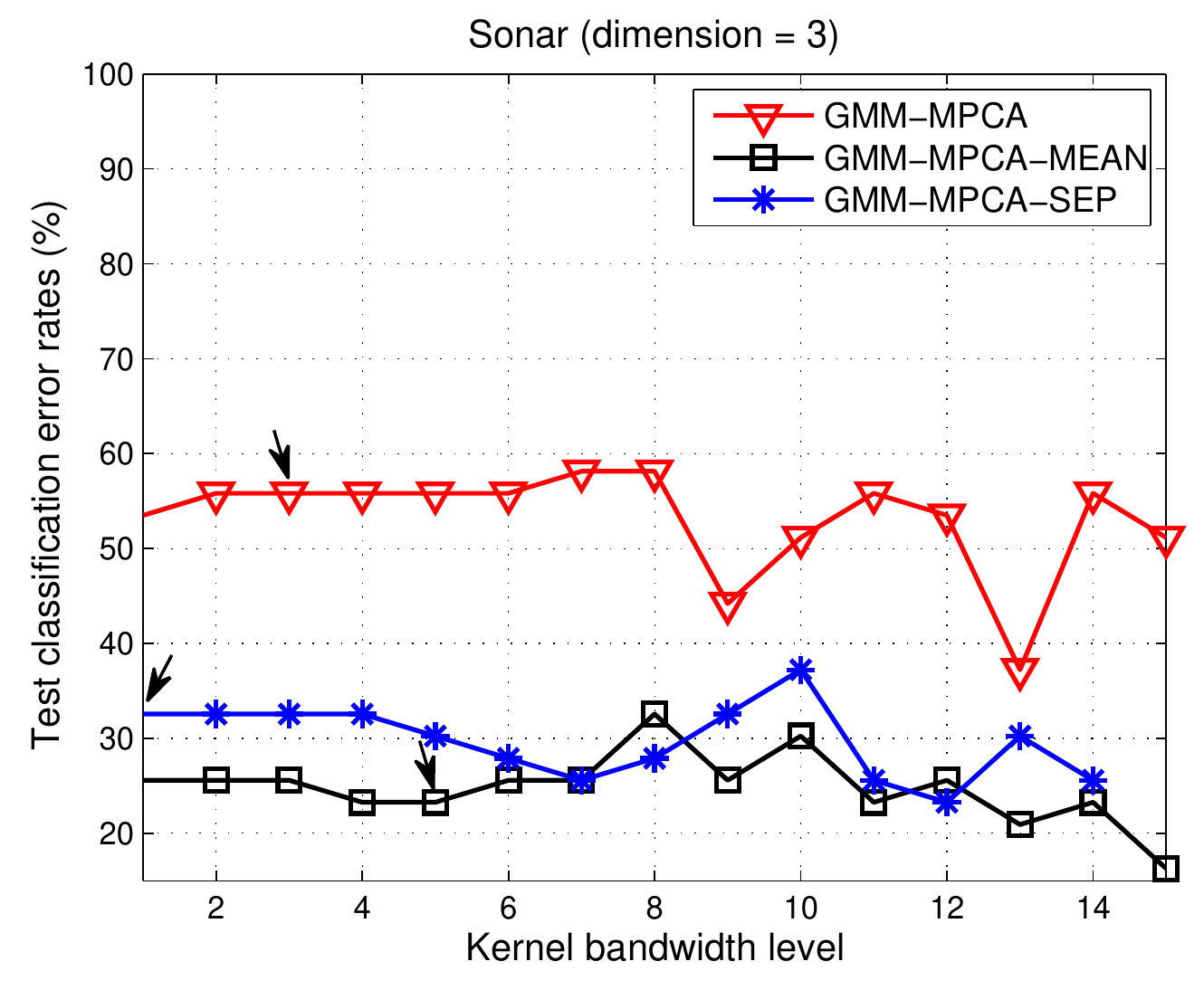,width=2.4in, origin=br,angle=0}
\epsfig{file=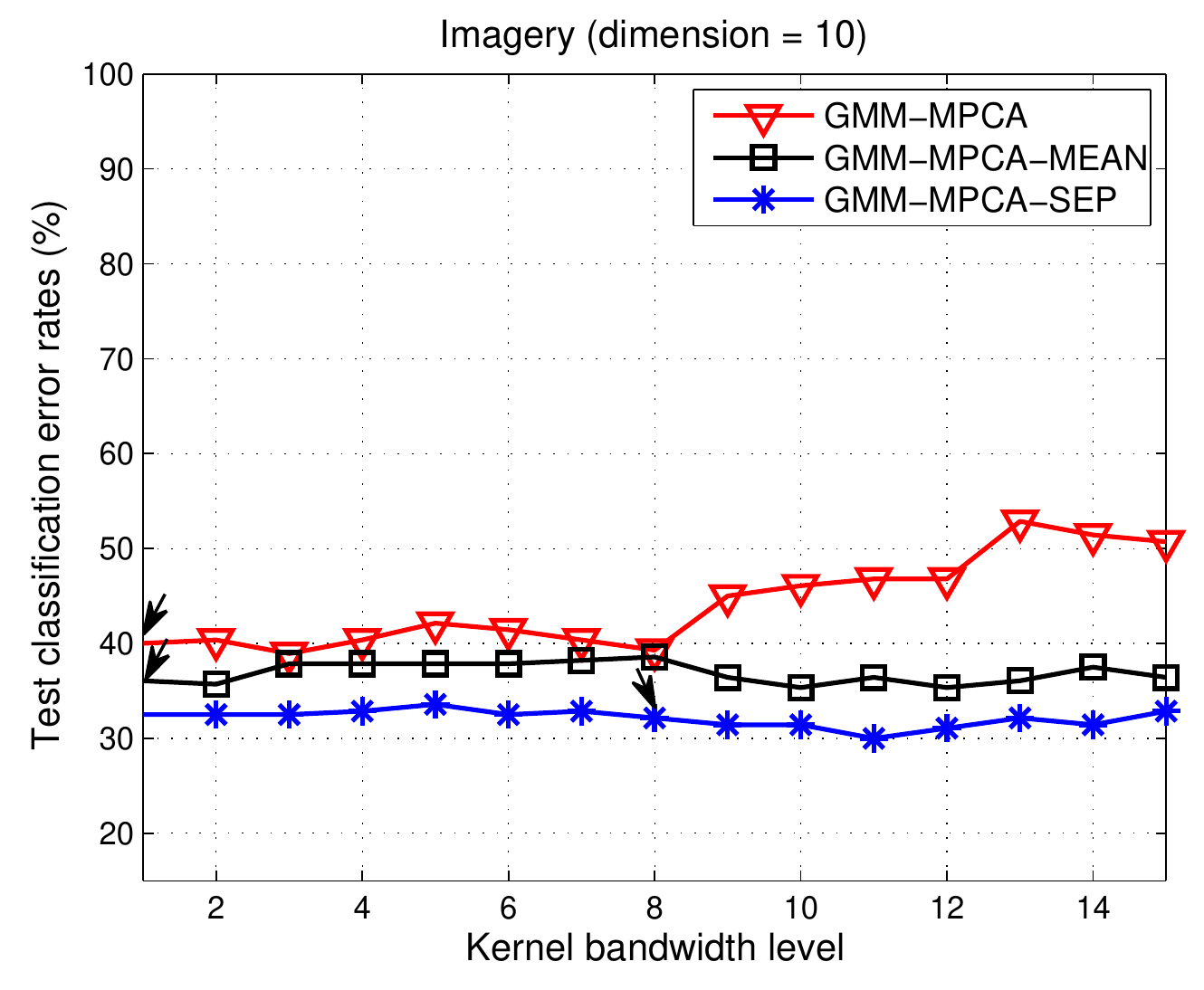,width=2.4in, origin=br,angle=0} \\
\end{tabular}
\caption{The test classification error rates at different levels of kernel bandwidth}
\label{fig:test}
\end{figure}

\begin{figure}[t]
\centering
\begin{tabular}{cc}
\epsfig{file=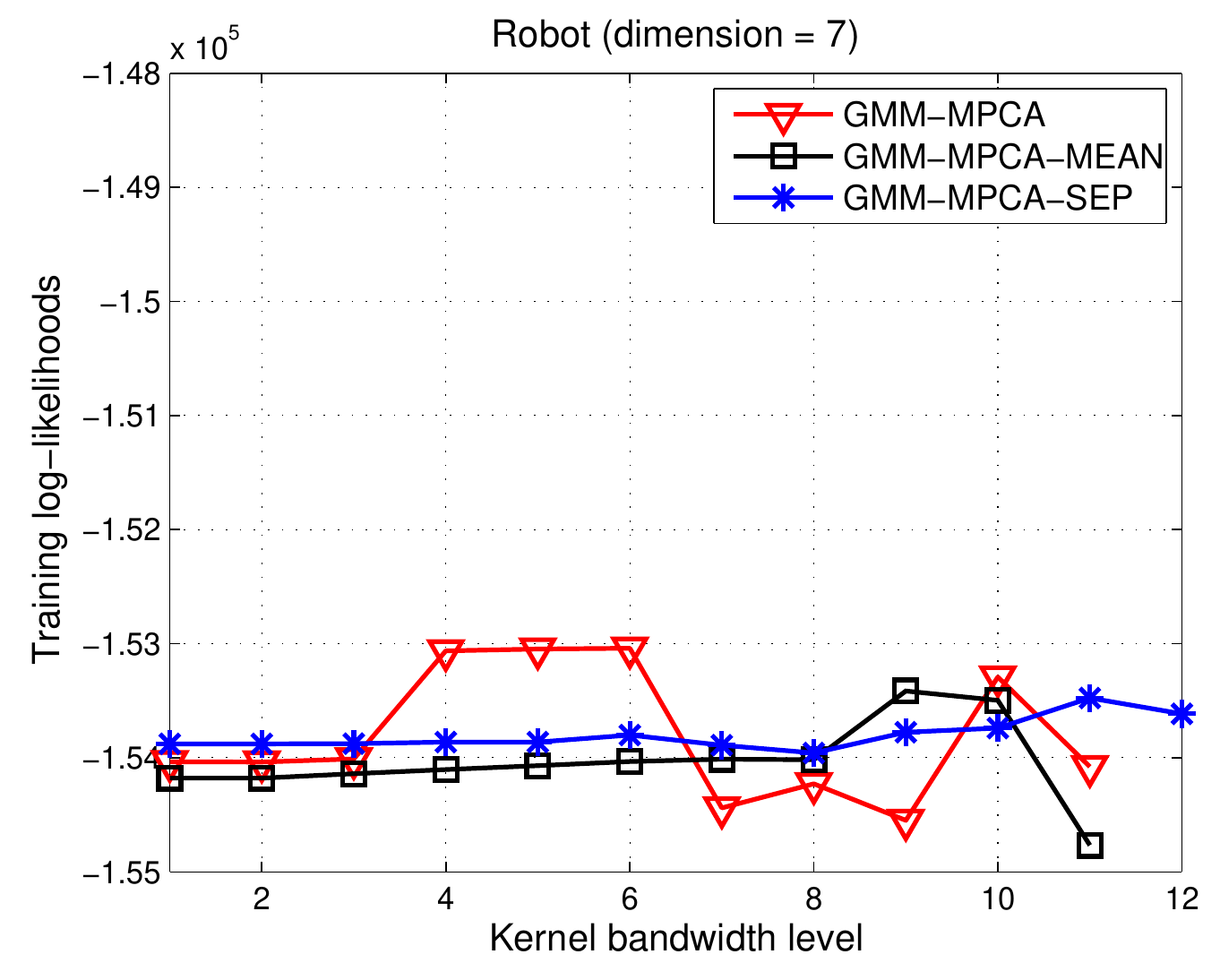,width=2.4in}
\epsfig{file=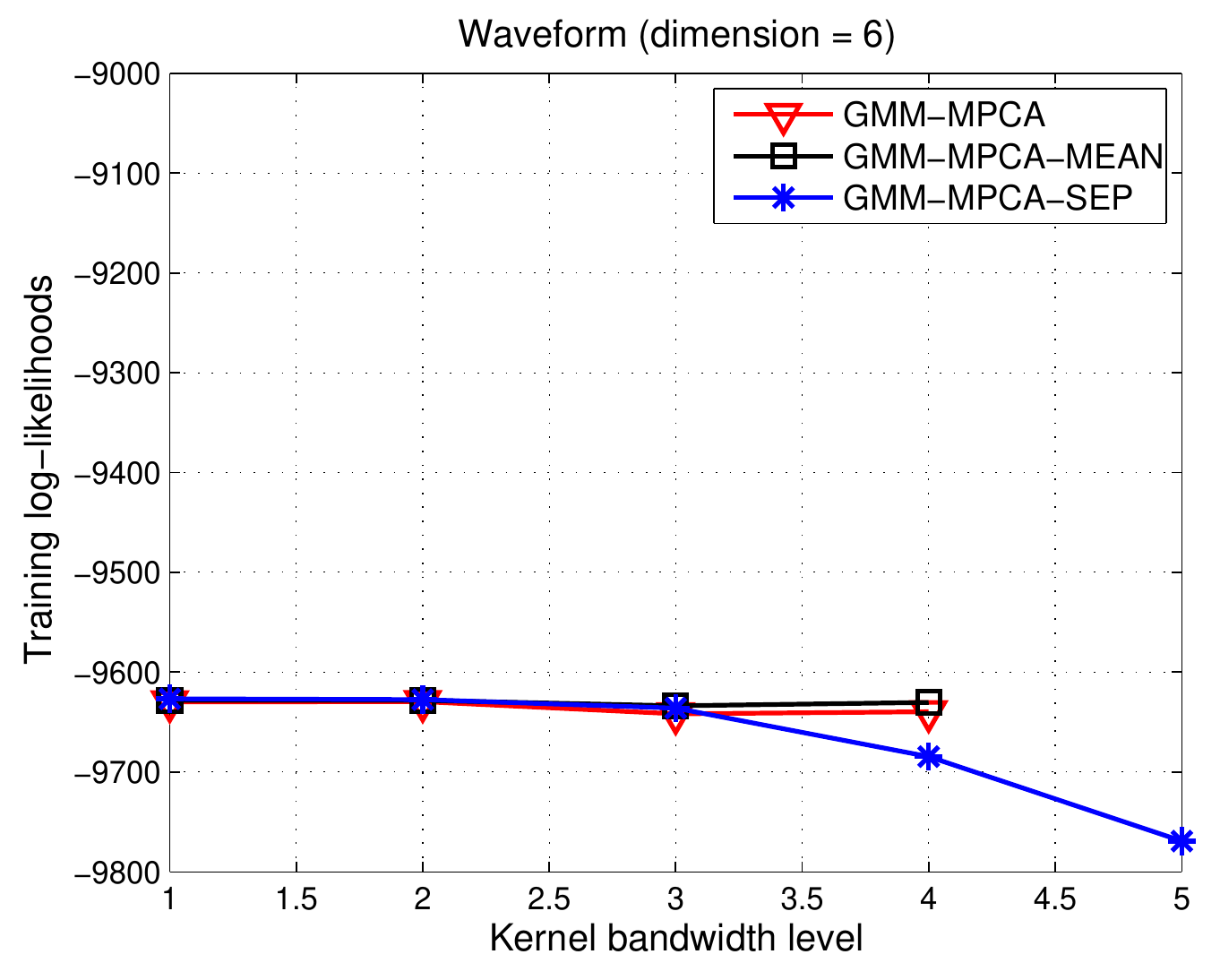,width=2.4in} \\
\epsfig{file=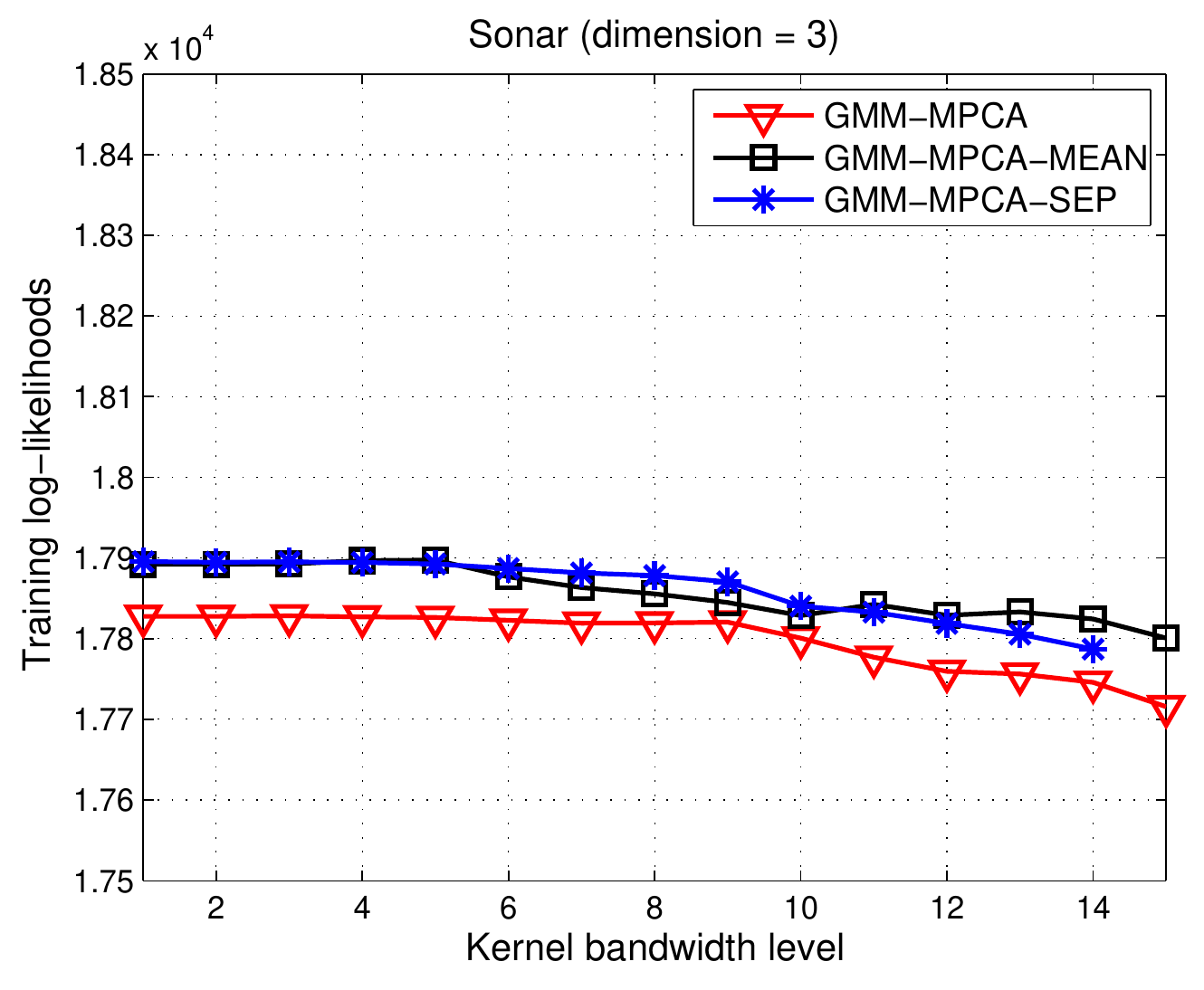,width=2.4in} 
\epsfig{file=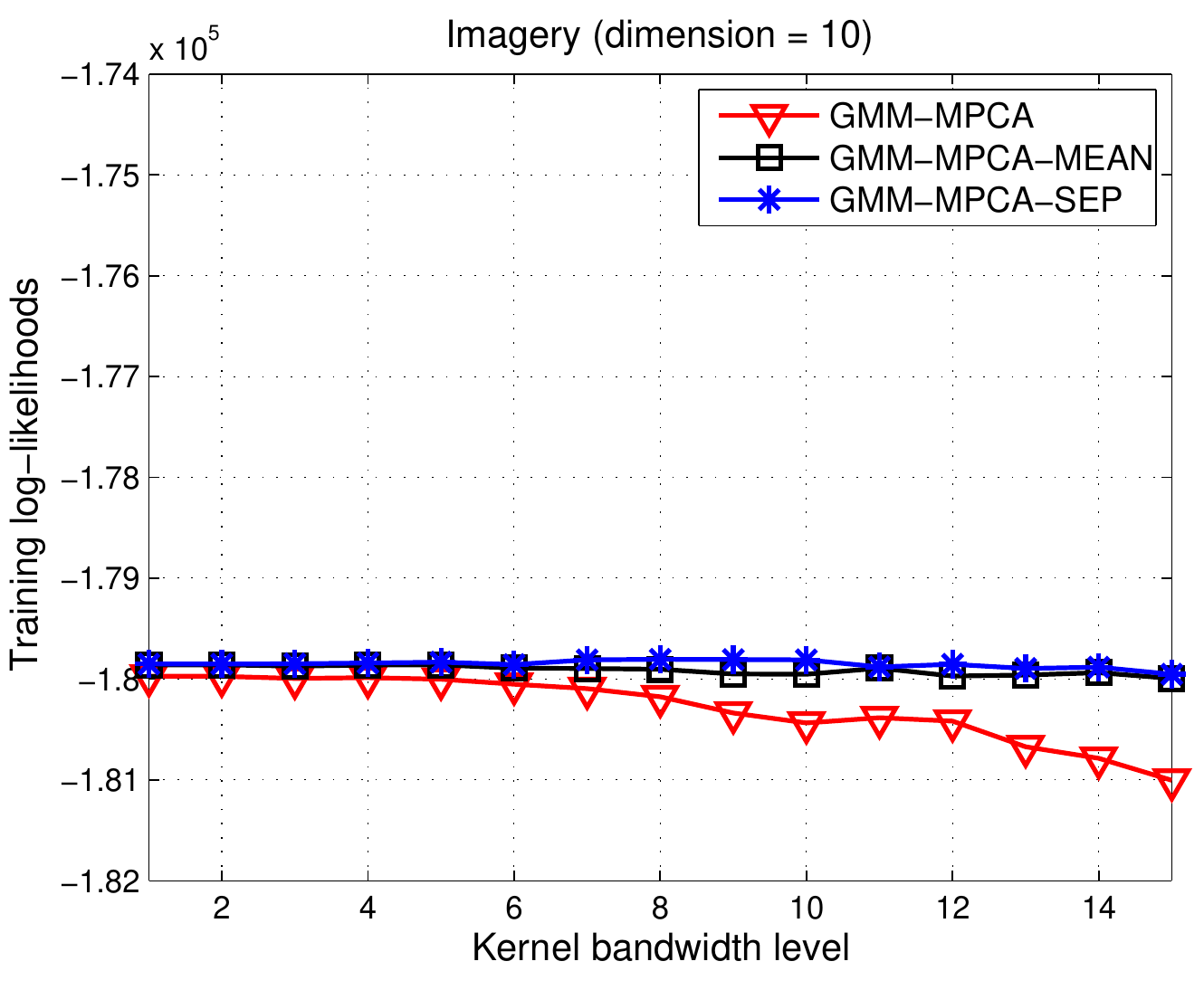,width=2.4in} \\
\end{tabular}
\caption{Training log-likelihoods at different kernel bandwidth levels}
\label{fig:training}
\end{figure}

Figure~\ref{fig:test} shows the test classification error rates at different levels of kernel 
bandwidth for several data sets (from one fold in the previous five-fold cross validation setup), when the number of mixture components for each class is set to three. 
The error rates are close to each other at the first few levels. 
As the kernel bandwidth increases, the error rates start to change. 
Except for the waveform, on which the error rates of GMM-MPCA and GMM-MPCA-MEAN are very close, 
for the other data sets in Figure~\ref{fig:test}, the error rate of GMM-MPCA-MEAN at each bandwidth level is 
lower than that of GMM-MPCA. Similarly, at each kernel bandwidth level, the error rate of GMM-MPCA-SEP is also lower 
than that of GMM-MPCA, except for the robot data. We also show the training log-likelihoods of these methods
with respect to different kernel bandwidth levels in Figure~\ref{fig:training}. The training log-likelihoods are also stable at 
the first few levels and start to fluctuate as the bandwidth increases. This is due to the possible big change in 
subspaces under large kernel bandwidths. 

In our model selection strategy, the subspace which results in the maximum log likelihood of the training model is selected and then we 
apply the model under the constraint of that specific subspace to classify the test data. In Figure~\ref{fig:test}, the test error rate
of the model which has the largest training likelihood is indicated by an arrow. As we can see, for each method,
this error rate is mostly ranked in the middle among all the error rates at different levels of bandwidth, which indicate that our model selection strategy helps find a reasonable training model.

\begin{table}[h!t!p!]
\tiny
\caption{Mean closeness of subspaces by MPCA and MPCA-MEAN at different levels of kernel bandwidth}
\vspace{-0.3cm}
\begin{center}
\subtable[Sonar data]{
		\begin{tabular}{|c||c|c|c|c|c|c|c|}
		\hline
Bandwidth level & 2 & 4 & 6 & 8 & 10 & 12 & 14 \\
\hline
Num of modes & 158	&	144		& 114		& 86	& 	60	& 	15	& 	8		\\
\hline
$d=2$ & (2.00, 2.00)	& (2.00, 2.00)	& (2.00, 2.00)	& (1.99, 1.99)	& (1.98, 1.98)& (1.77, 1.88)	& (1.43, 1.80) \\
$d=4$ & (4.00, 4.00)	& (4.00, 4.00)	& (3.99, 4.00)	& (3.98, 3.99)	& (3.84, 3.94)	& (2.63, 3.08)	& (2.09, 2.96) \\
$d=6$ & (6.00, 6.00)	& (5.99, 5.99)	& (5.95, 5.99)	& (5.91, 5.88)	& (5.41, 5.47)	& (4.48, 4.64)	& (3.48, 4.13) \\
$d=8$ & (8.00, 8.00)	& (8.00, 8.00)	& (7.97, 7.97)	& (7.86, 7.89)	& (6.98, 7.00)	& (6.20, 6.40)	& (4.29, 5.18) \\
\hline
\end{tabular}
}
\subtable[Imagery data]{
		\begin{tabular}{|c||c|c|c|c|c|c|c|}
		\hline
Bandwidth level & 2	& 4	& 6	& 8	 & 10	& 12	& 14 \\
\hline
Num of modes & 1109	& 746	& 343	& 144	& 60 & 35	& 14 \\
\hline
$d=6$ & (6.00, 6.00)	& (5.97, 5.99) & (5.32, 5.86)& (5.34, 5.61)& (5.21, 5.32)& (5.02, 5.32)& (3.63, 5.15)\\
$d=8$ & (8.00, 8.000)	& (7.96, 7.96) & (7.54, 7.89)& (7.31, 7.76)& (6.82, 7.43)& (6.27, 6.85)& (4.83, 6.61)\\
$d=10$ &(10.00, 10.00)  & (9.80, 9.52) & (9.56, 9.48)& (9.25, 9.23)& (8.45, 9.01)& (7.71, 8.42)& (5.86, 7.55)\\
$d=12$ &(12.00, 12.00)& (11.96, 11.96)	& (11.48, 11.50)	& (10.29, 10.89)	& (10.06, 10.33)& (9.49, 9.87)	& (7.42, 8.98)\\
\hline
\end{tabular}
}
\end{center}
\label{table:sensitivity1}
\end{table}


\subsection{Clustering}\label{vis:clustering}
We present the clustering results of GMM-MPCA and MDA-RR on several 
simulated data sets and visualize the results in a low-dimensional subspace. 
The previous model selection criteria is also used in clustering. 
After fitting a subspace constrained Gaussian mixture
model, all the data points having the highest posterior probability
belonging to a particular component form one cluster. We outline the 
data simulation process as follows.

The data is generated from some existing subspace constrained
model. Specifically, we take the training set of the
imagery data 
from one fold in the previous five-fold cross validation setup
and estimate its distribution by
fitting a mixture model via GMM-MPCA. We will obtain five estimated
component means which are ensured to be constrained in a two dimensional subspace. 
A set of $200$ samples are randomly drawn from a multivariate Gaussian
distribution with the previously estimated component means as the
sample means. A common identity covariance is assumed for the Gaussian
multivariate distributions. We generate five sets of samples in this way, forming a data set of $1000$ samples. We scale the component means by 
different factors so that the data have different levels of dispersion among the clusters. The lower the dispersion, 
the more difficult the clustering task. 
Specifically, the scaling factor
is set to be 0.125, 0.150, and 0.250, respectively, generating three simulated data with low, middle and high level 
dispersion between clusters. 

\begin{figure}[t!]
\centering
\begin{tabular}{ccc}
\epsfig{file=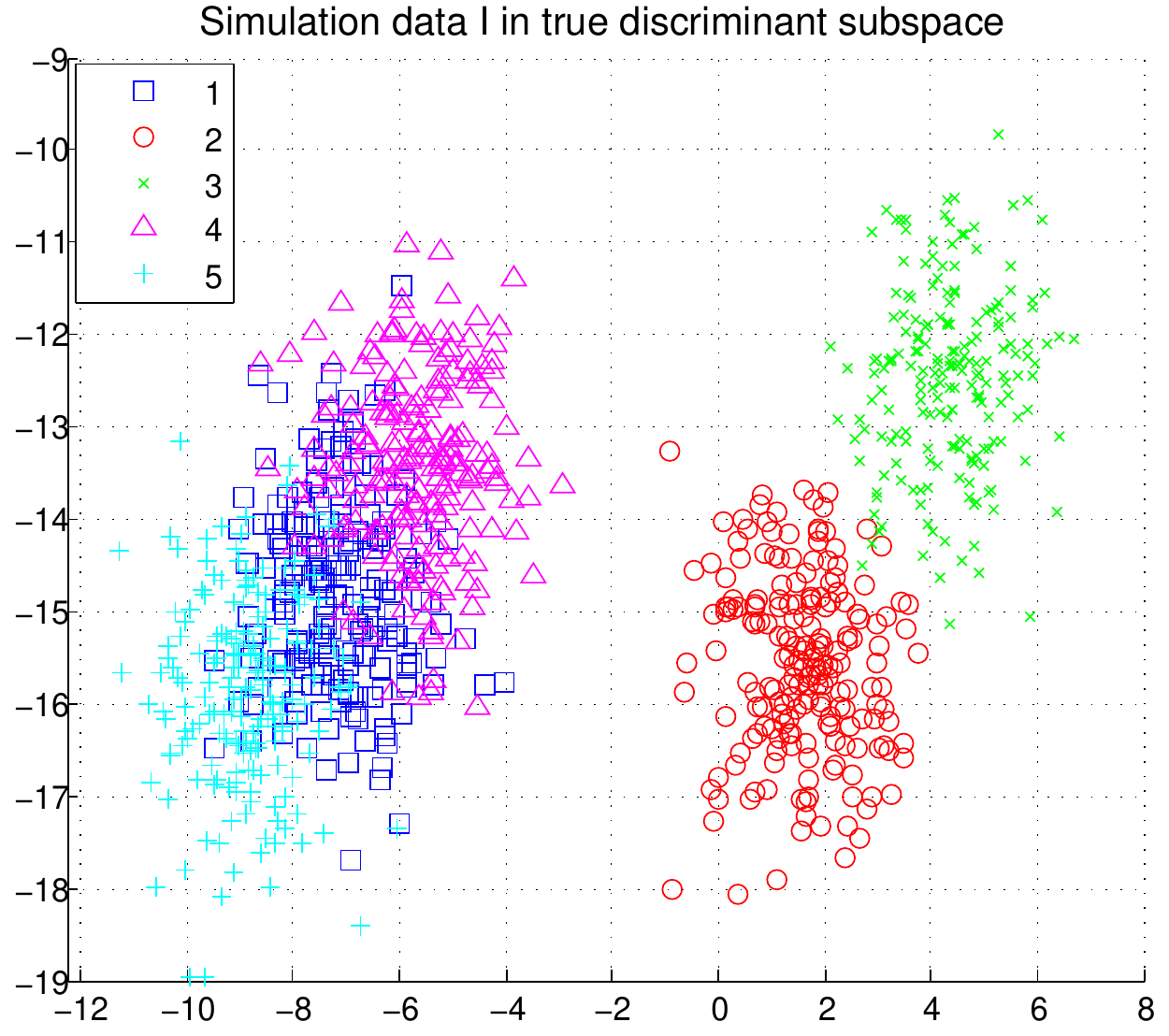,width=1.92in} &
\epsfig{file=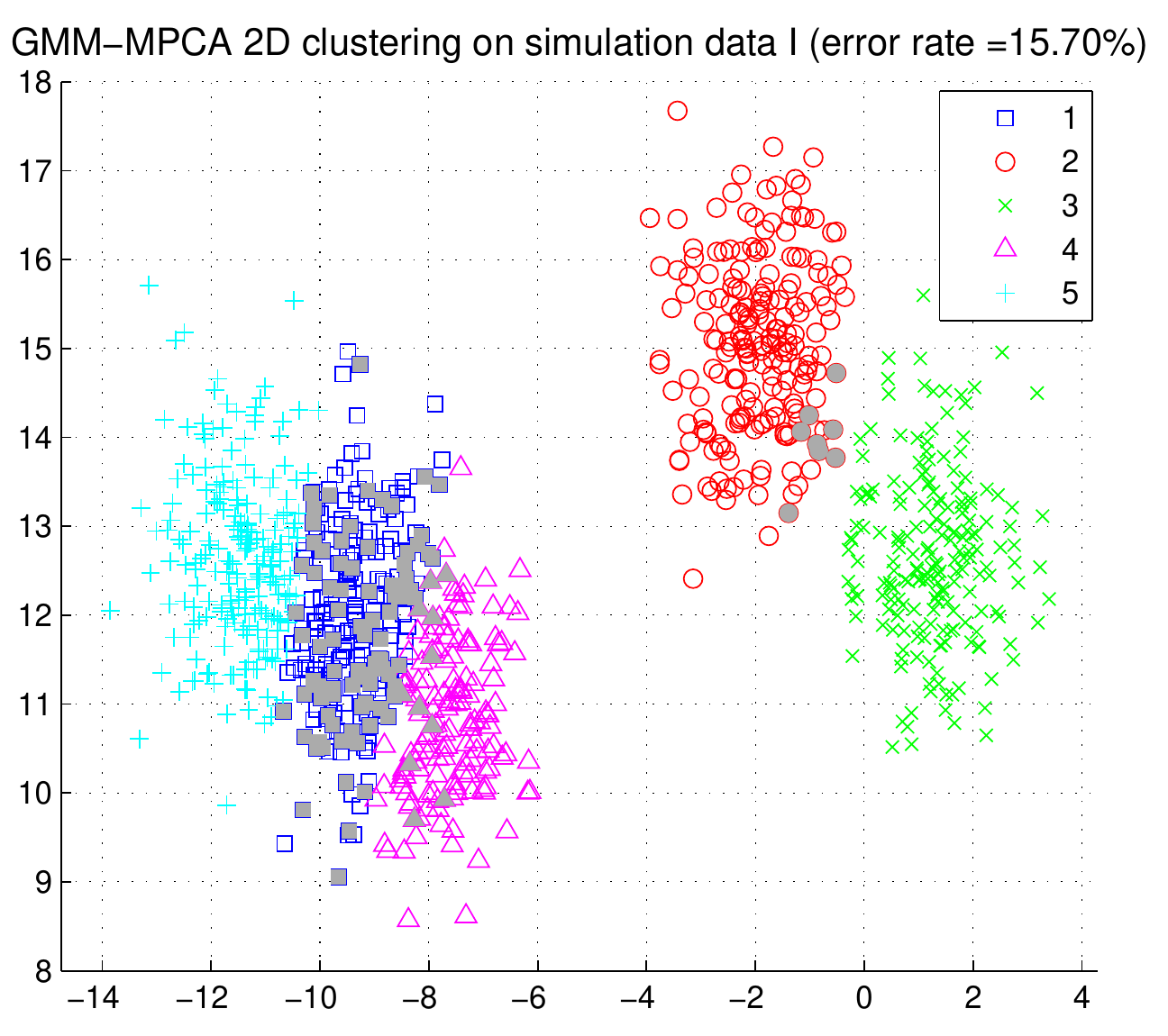,width=1.95in}  & 
\epsfig{file=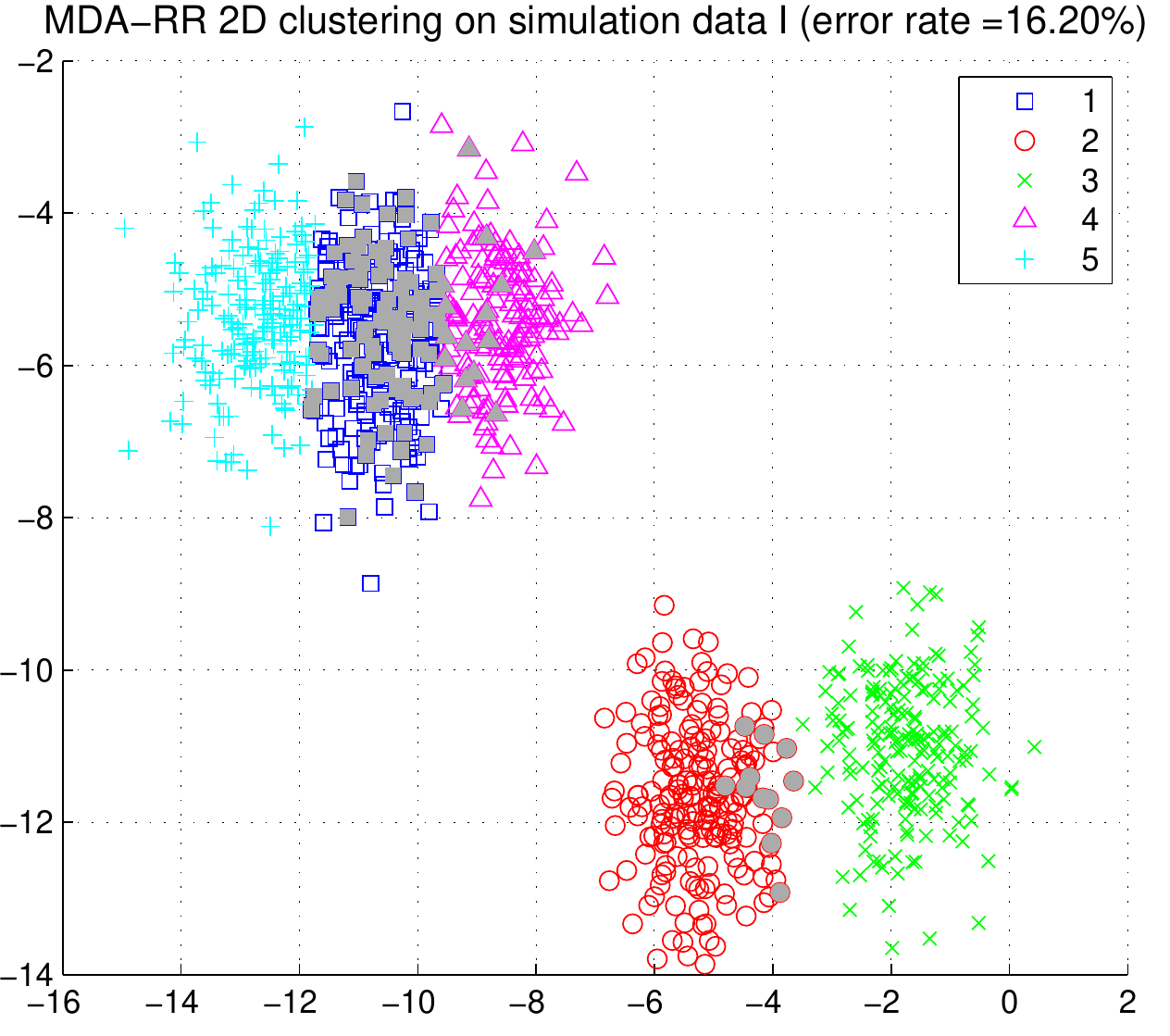,width=1.90in} \\
\epsfig{file=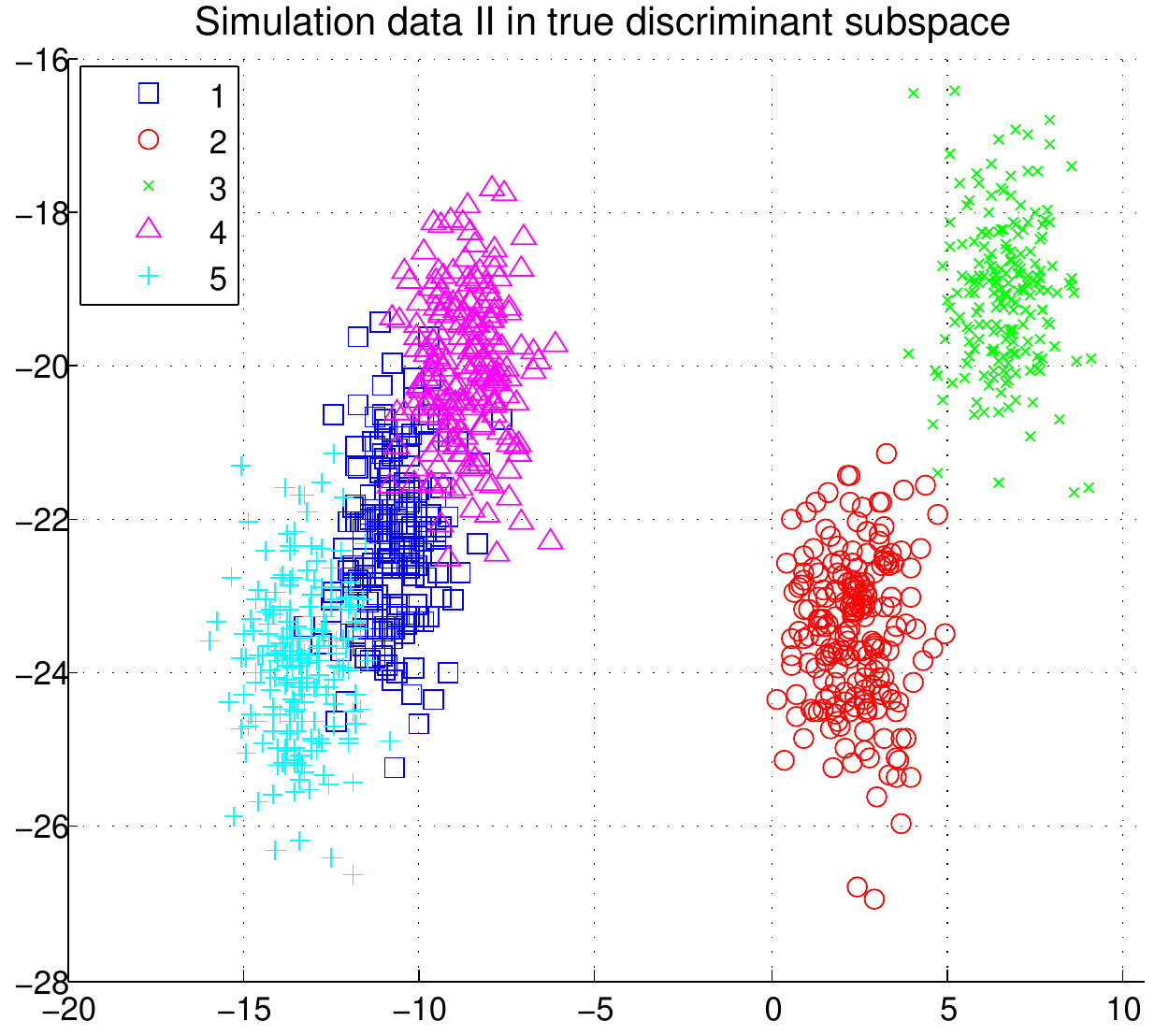,width=1.92in} &
\epsfig{file=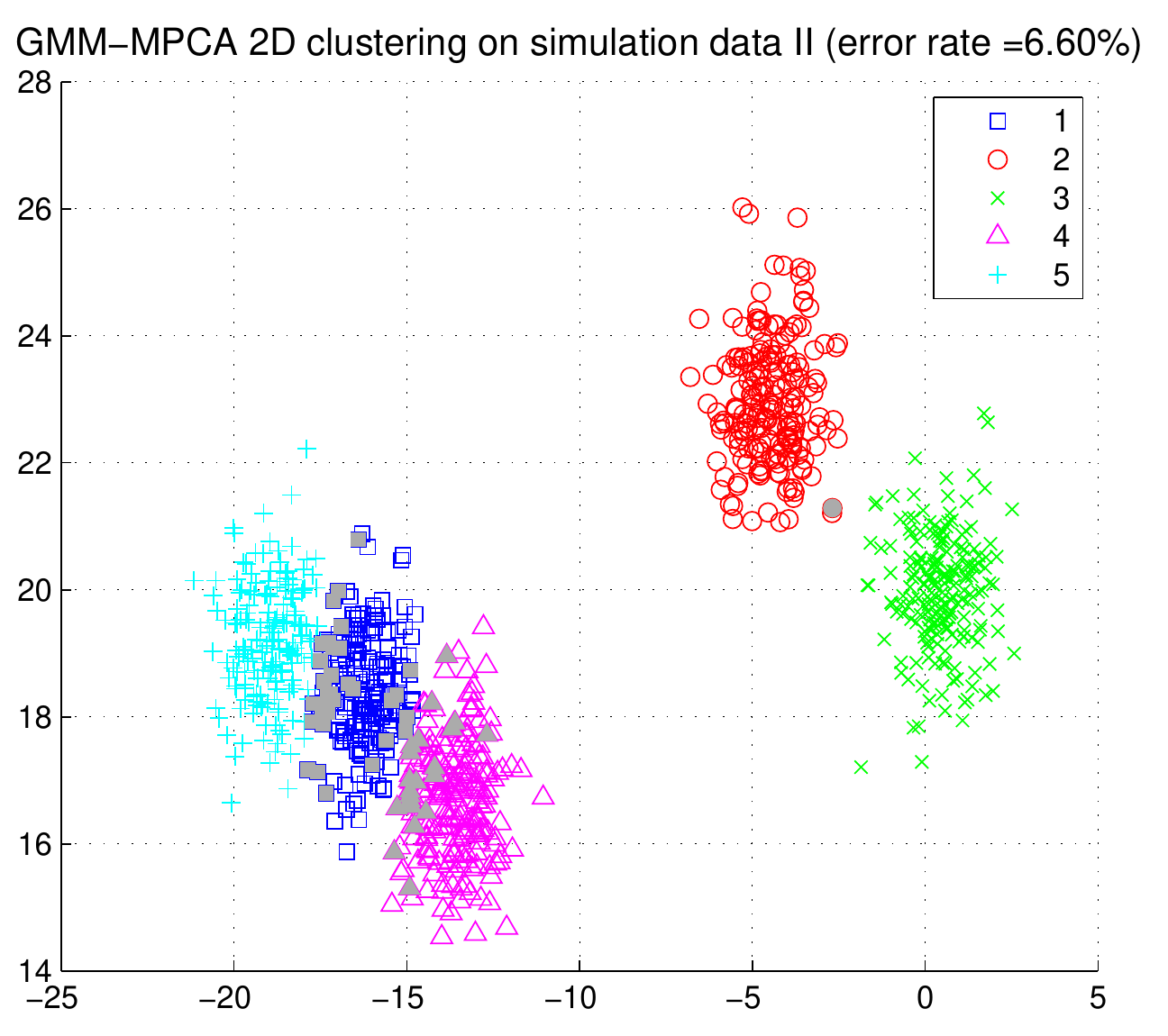,width=1.95in}  & 
\epsfig{file=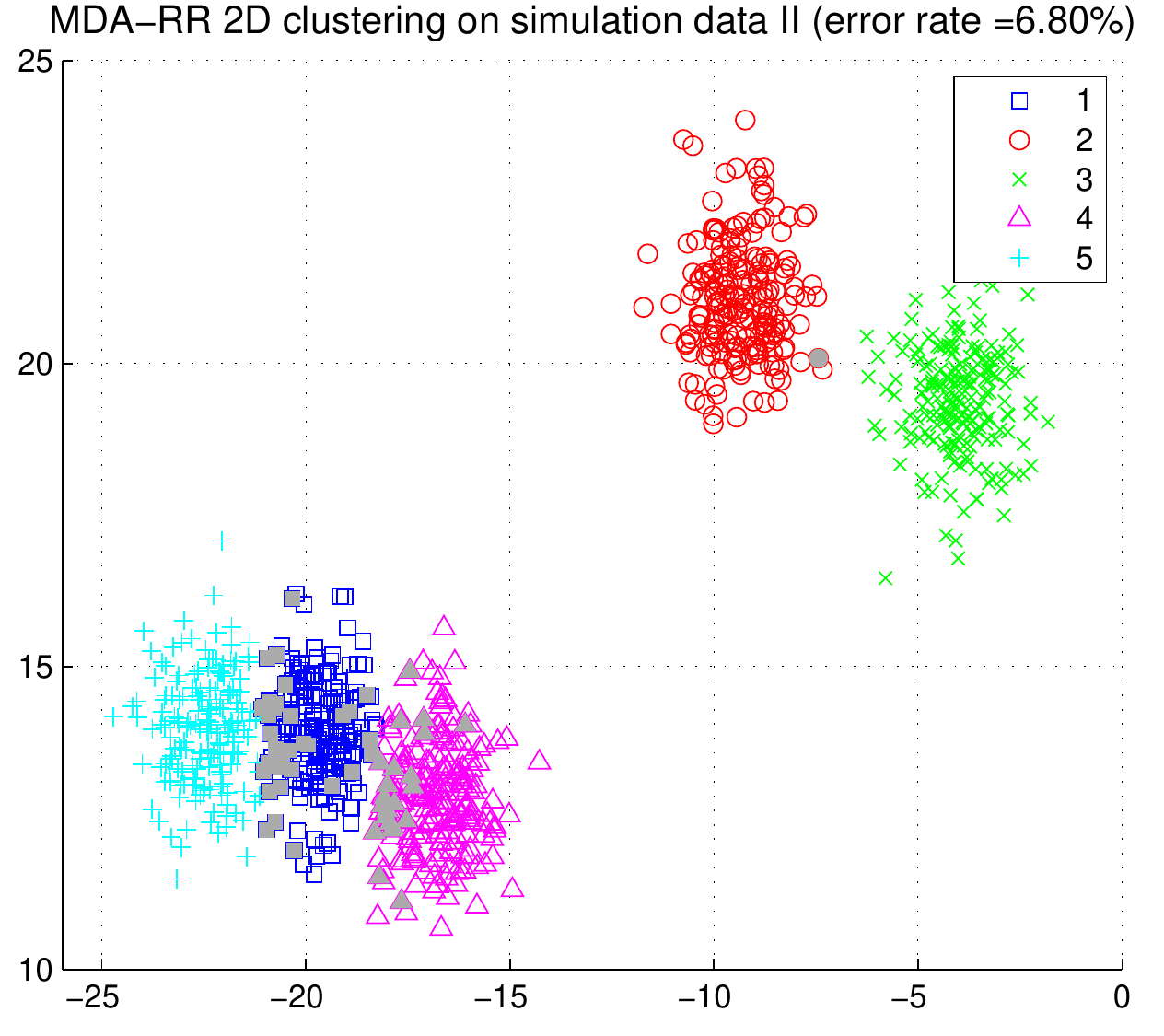,width=1.90in} \\
\epsfig{file=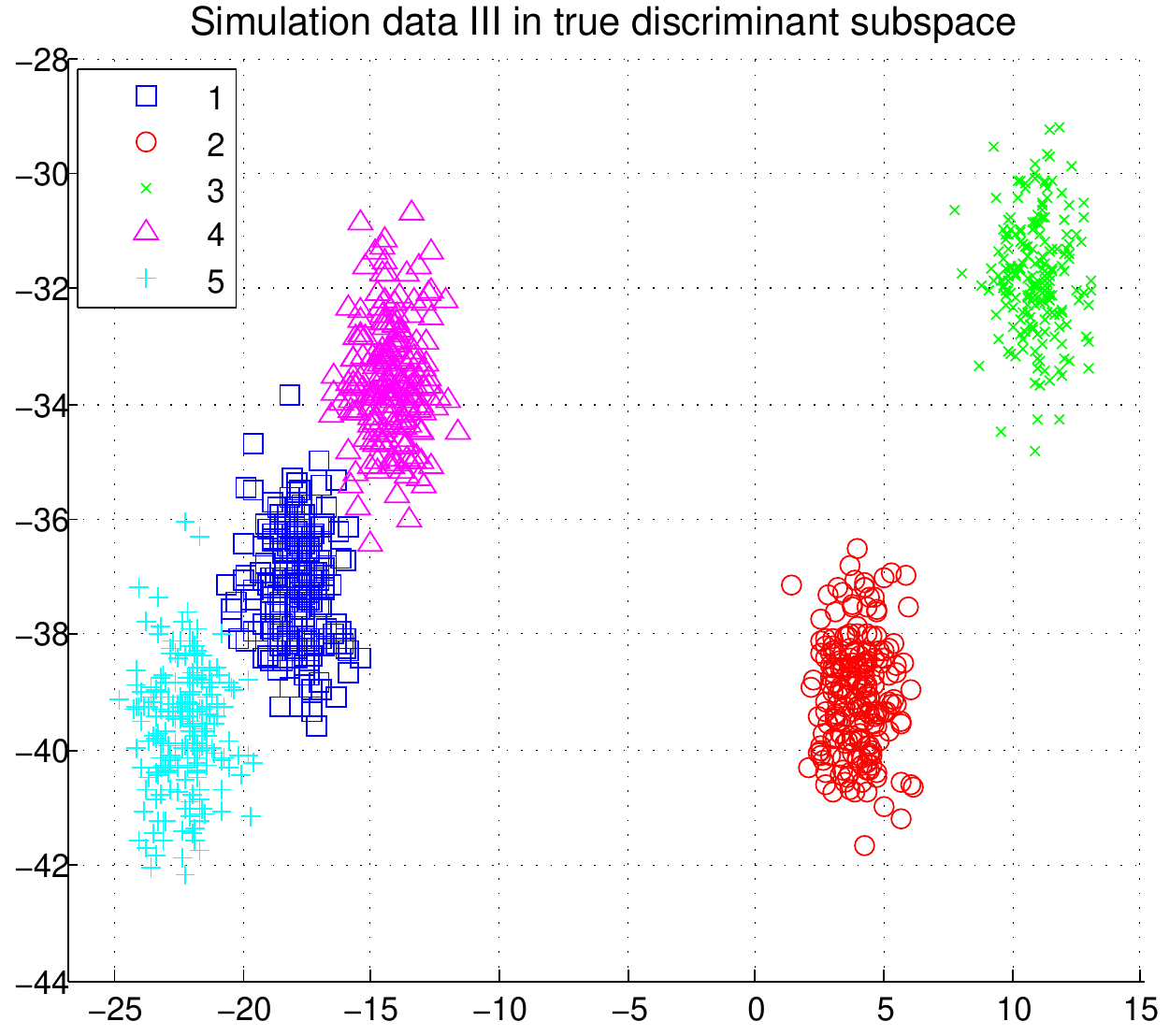,width=1.92in} &
\epsfig{file=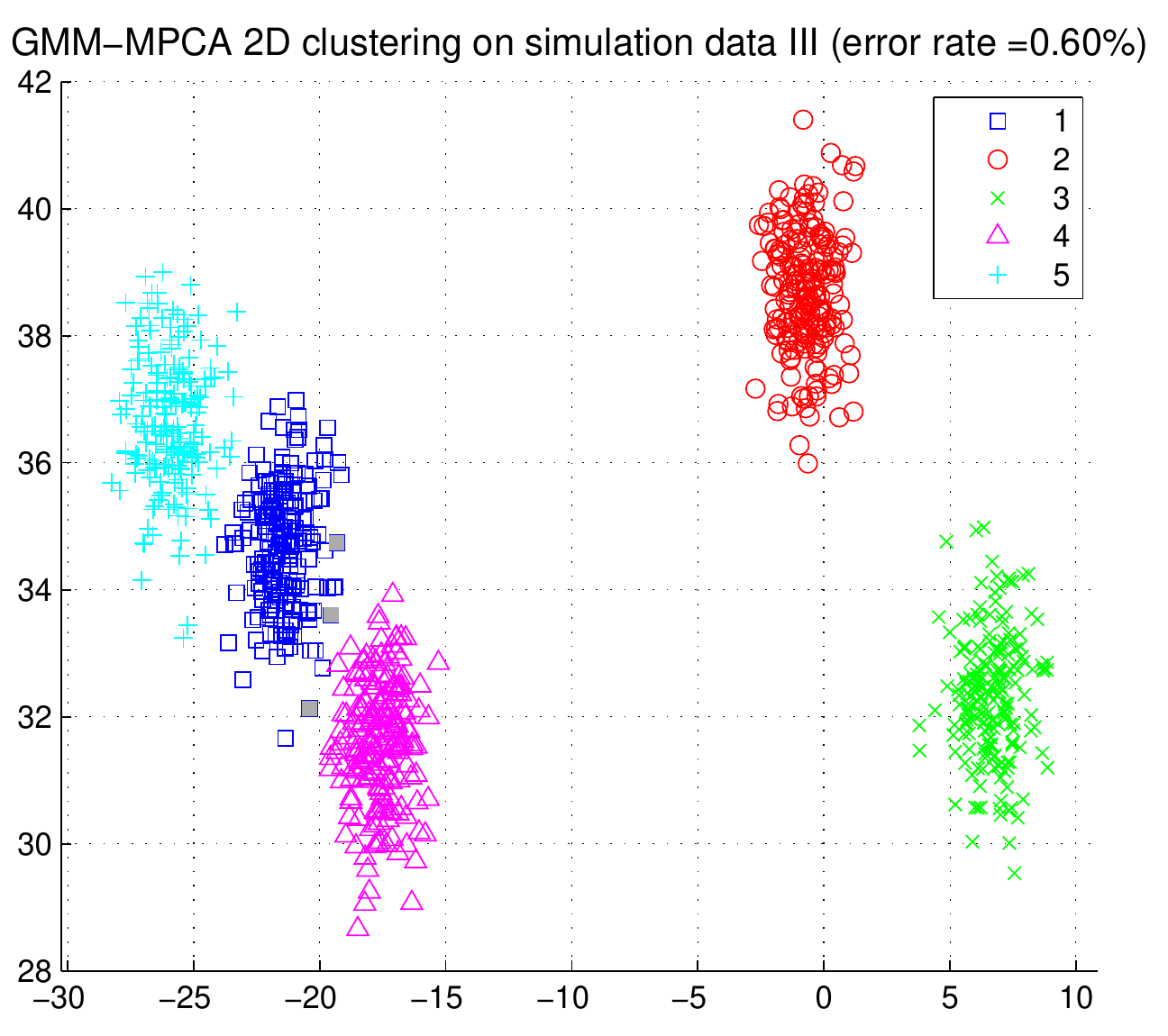,width=1.95in}  & 
\epsfig{file=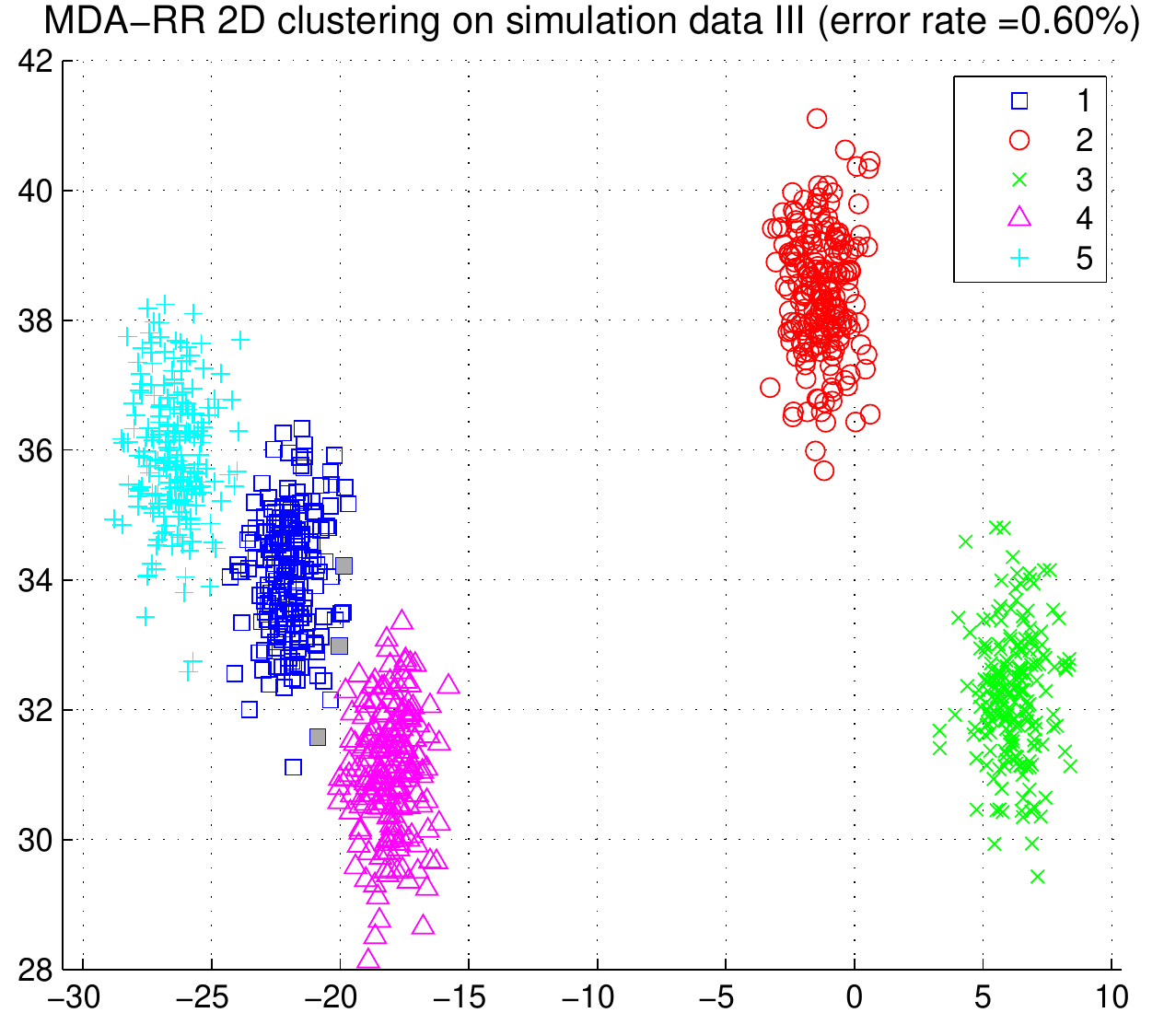,width=1.90in}  \\
(a) & (b) & (c)\\
\end{tabular}
\caption{Two-dimensional plot for the clustering of synthetic data,
color-coding the clusters. (a) Projections onto the
two-dimensional true discriminant subspace, with true cluster labels. 
(b), (c) Projections onto the
two-dimensional discriminant subspace by GMM-MPCA and MDA-RR respectively, with predicted cluster labels. 
}
\label{fig:visclustering}
\end{figure}

Figure~\ref{fig:visclustering} shows the clustering results of 
three simulated data sets by GMM-MPCA and MDM-RR, in two-dimensional plots,
color-coding the clusters. The data projected onto the true discriminant subspace with true cluster labels are shown in 
Figure~\ref{fig:visclustering}(a). In addition, 
Figure~\ref{fig:visclustering}(b) and Figure~\ref{fig:visclustering}(c)
show the data projected onto the two-dimensional discriminant subspaces
by GMM-MPCA and MDA-RR. For all the simulated data sets, both
GMM-MPCA and MDA-RR can effectively reveal the clustering structure in a low-dimensional subspace. To evaluate their clustering 
performance, we compute the clustering accuracy by comparing their predicted and true clustering labels. Suppose the true 
cluster label of data point $\boldsymbol{x}_i$ is $t_i$ and the predicted cluster label is $p_i$, the clustering error rate is calculated as 
$1-\sum_{i=1}^nI(t_i,map(p_i))/n$, where $n$ is the total number of data points, $I(x,y)$ is an indicator function that is equal to one if $x=y$ otherwise zero, and $map(p_i)$ is a permutation function which maps the predicted label to an equivalent label in the data set. Specifically, we use the 
Kuhn-Munkres algorithm to find the best matching~\citep{Lovasz86}. The clustering error rates are listed in the titles above the plots in Figure~\ref{fig:visclustering}. The mis-clustered data points are in gray. When the dispersion between clusters is low or middle, the clustering error rates of GMM-MPCA are smaller than those of MDA-RR. 
When the dispersion is high, the task becomes relatively easy and the clustering accuracy of these two methods are the same. In Table~\ref{table:closeness}, we also show the closeness between the true discriminant subspace and the discriminant subspaces found by GMM-MPCA and MDA-RR. 
Comparing with MDA-RR, for all the three data sets, the closeness between the subspace by GMM-MPCA and the true subspace are smaller. 
\begin{table}
\caption{Closeness between subspaces in clustering with different dispersions}
\scriptsize
\centering
\begin{tabular}{|c || c c c|} 
\hline
Closeness & low & middle & high \\
\hline
GMM-MPCA & 1.769 & 1.820 & 1.881   \\
MDA-RR &  1.552 & 1.760 & 1.866 \\
\hline 
\end{tabular}
\label{table:closeness} 
\end{table}

\section{Conclusion} \label{sec:conclude}
In this paper, we propose a Gaussian mixture model with the component means 
constrained in a pre-selected subspace. We prove that the modes, the component means
of a Gaussian mixture, and the class means all  
lie in the same constrained subspace. Several approaches to  
finding the subspace are proposed 
by applying weighted PCA to the modes, class means, or 
a union set of modes and class means. The constrained method results in a dimension reduction property, which  
allows us to view the classification or clustering structure of the data in a lower
dimensional space. An EM-type algorithm is derived to estimate the model, given any constrained subspace. In addition, 
the Gaussian mixture model with the component means constrained by separate parallel subspace for 
each class is investigated. Although reduced rank MDA is a competitive classification method by constraining the class means to an optimal discriminant 
subspace within each EM iteration, experiments on several real data sets of moderate to high dimensions
show that when the dimension of the discriminant subspace is very low, it is often outperformed by our proposed method with a simple technique of spanning the constrained subspace using only class means.

We select the constrained subspace which has the largest training likelihood among a sequence of
subspaces resulting from different kernel bandwidths. If the number
of candidate subspaces is large, it may be desired to 
narrow down the search by incorporating some prior knowledge. 
For instance, the proposed method may have a potential 
in visualization when users already know that only a certain dimensions of the data matter
for classification or clustering, i.e., a constrained subspace can be obtained beforehand. 
Finally, we expect this subspace constrained method can be extended to other parametric
mixtures, for instance, mixture of Poisson for discrete data. 

{

}

\clearpage
\begin{center}
{\bf SUPPLEMENTAL MATERIALS LIST}
\end{center}
{\bf Appendix to Gaussian Mixture Models with Component
Means Constrained in Pre-selected Subspaces:} In Appendix A, we prove Theorem 3.1. The proof of Theorem 3.2 
is provided in Appendix B. The derivation of of $\mu_{kr}$ in GEM is in Appendix C and 
reduced rank mixture discriminant analysis is in Appendix D.
\vspace{0.5cm}


\clearpage
\begin{center}
{\large\bf SUPPLEMENTAL MATERIALS}
\end{center}
\section*{Appendix A}\label{appendix1}
We prove Theorem 3.1. Consider a mixture of Gaussians with 
a common covariance matrix $\boldsymbol{\Sigma}$ shared across all the components as in (2):
\[f(\boldsymbol{X}=\boldsymbol{x})=\sum_{r=1}^{R}\pi_{r} \phi(\boldsymbol{x}|\boldsymbol{\mu}_{r},\boldsymbol{\Sigma})\;.\]
Once
$\boldsymbol{\Sigma}$ is identified, a linear transform (a
``whitening'' operation) can be applied to $\boldsymbol{X}$ so that the
transformed data follow a mixture with component-wise diagonal
covariance, more specifically, the identity matrix $\boldsymbol{I}$.  Assume
$\boldsymbol{\Sigma}$ is non-singular and hence positive definite, we
can find the matrix square root of $\boldsymbol{\Sigma}$, that is,
$\boldsymbol{\Sigma}=(\boldsymbol{\Sigma}^{\frac{1}{2}})^t
\boldsymbol{\Sigma}^{\frac{1}{2}}$.  If the eigen decomposition of
$\boldsymbol{\Sigma}$ is
$\boldsymbol{\Sigma}=V_{\boldsymbol{\Sigma}}D_{\boldsymbol{\Sigma}}
V_{\boldsymbol{\Sigma}}^{t}$, then,
$\boldsymbol{\Sigma}^{\frac{1}{2}}=D_{\boldsymbol{\Sigma}}^{\frac{1}{2}}V_{\boldsymbol{\Sigma}}^{t}$.
Let
$W=((\boldsymbol{\Sigma}^{\frac{1}{2}})^t)^{-1}$ and
$\boldsymbol{Z}=W\boldsymbol{X}$.  The density of $\boldsymbol{Z}$ is
$g(\boldsymbol{Z}=\boldsymbol{z})=\sum_{r=1}^{R}\pi_{r} \phi(\boldsymbol{z}|W
\boldsymbol{\mu}_{r}, \boldsymbol{I})$. Any mode of $g(\boldsymbol{z})$
corresponds to a mode of $f(x)$ and vice versa.  Hence, without loss
of generality, we can assume $\boldsymbol{\Sigma}=\boldsymbol{I}$.

Another linear transform on $\boldsymbol{Z}$ can be performed using the
orthonormal basis $V=\boldsymbol{\nu}\cup\boldsymbol{\nu}^{\perp}=
\{\boldsymbol{v}_1, ..., \boldsymbol{v}_p\}$, where 
$\boldsymbol{\nu}=\{\boldsymbol{v}_{q+1}, ..., \boldsymbol{v}_p\}$ is the constrained subspace where $\boldsymbol{\mu}_{kr}$'s reside, 
and $\boldsymbol{\nu}^{\perp}=\{\boldsymbol{v}_1, ..., \boldsymbol{v}_q\}$ is the corresponding null subspace, as defined in 
Section 3. Suppose $\widetilde{\boldsymbol{Z}}=\textbf{Proj}_{V}^{\boldsymbol{Z}}$. 
For the transformed data
$\widetilde{\boldsymbol{z}}$, the covariance matrix is still $\boldsymbol{I}$. Again, there is a one-to-one correspondence (via the orthonormal
linear transform) between the modes of $g_k(\widetilde{\boldsymbol{z}})$
and the modes of $g_k(\boldsymbol{z})$. The density of
$\widetilde{\boldsymbol{z}}$ is
\[
g(\widetilde{\boldsymbol{Z}}=\widetilde{\boldsymbol{z}})=\sum_{r=1}^{R}\pi_{r}\phi(\widetilde{\boldsymbol{z}}|\boldsymbol{\theta}_{r},\boldsymbol {I})\;,
\]
where $\boldsymbol{\theta}_{r}$ is the projection of $W\boldsymbol{\mu}_{r}$
onto the orthonormal basis $V$, i.e., $\boldsymbol{\theta}_{kr}=\textbf{Proj}_{V}^{W\boldsymbol{\mu}_{r}}$.  Split
$\boldsymbol{\theta}_{r}$ into two parts, $\boldsymbol{\theta}_{r,1}$
being the first $q$ dimensions of $\boldsymbol{\theta}_{r}$ and
$\boldsymbol{\theta}_{r,2}$ being the last $p-q$ dimensions.  Since
the projections of $\boldsymbol{\mu}_{r}$'s onto 
the null subspace $\boldsymbol{\nu}^{\perp}$ are the same, 
$\boldsymbol{\theta}_{r,1}$ are identical for all the components, which is hence denoted by $\boldsymbol{\theta}_{\cdot,1}$.
Also denote the first $q$ dimensions of $\widetilde{\boldsymbol{z}}$ by
$\widetilde{\boldsymbol{z}}^{(1)}$, and the last $p-q$ dimensions by
$\widetilde{\boldsymbol{z}}^{(2)}$.  We can write
$g(\widetilde{\boldsymbol{z}})$ as
\begin{eqnarray}
g(\widetilde{\boldsymbol{Z}}=\widetilde{\boldsymbol{z}})=\sum_{r=1}^{R}\pi_{r}\phi(\widetilde{\boldsymbol{z}}^{(1)}|\boldsymbol{\theta}_{\cdot,1},\boldsymbol{I}_q)
\phi(\widetilde{\boldsymbol{z}}^{(2)}|\boldsymbol{\theta}_{r, 2},\boldsymbol{I}_{p-q})\;. \nonumber
\label{eq:split}
\end{eqnarray}
where $\boldsymbol{I}_q$ indicates a $q\times q$ identity matrix. Since $g(\widetilde{\boldsymbol{z}})$ is a smooth function, 
its modes have zero first order derivatives. Note
\[
\frac{\partial g(\widetilde{\boldsymbol{z}})}{\partial
  \widetilde{\boldsymbol{z}}^{(1)}}=\frac{\partial \phi(\widetilde{\boldsymbol{z}}^{(1)}|\boldsymbol{\theta}_{\cdot, 1},\boldsymbol{I}_{q})}{\partial \widetilde{\boldsymbol{z}}^{(1)}} \sum_{r=1}^{R}
\pi_{r} \phi(\widetilde{\boldsymbol{z}}^{(2)}|\boldsymbol{\theta}_{r,2},\boldsymbol{I}_{p-q})\;,
\]
\[
\frac{\partial g(\widetilde{\boldsymbol{z}})}{\partial
  \widetilde{\boldsymbol{z}}^{(2)}}=\phi(\widetilde{\boldsymbol{z}}^{(1)}|\boldsymbol{\theta}_{\cdot, 1},\boldsymbol{I}_{q}) \sum_{r=1}^{R}
\pi_{r} \frac{\partial \phi(\widetilde{\boldsymbol{z}}^{(2)}|\boldsymbol{\theta}_{r,2},\boldsymbol{I}_{p-q})}{\partial
  \widetilde{\boldsymbol{z}}^{(2)}} \;.
\]
By setting the first partial derivative to zero and using the fact
$\sum_{r=1}^{R}
\pi_{r} \phi(\widetilde{\boldsymbol{z}}^{(2)}|\boldsymbol{\theta}_{r,2},\boldsymbol{I}_{p-q})>0$, we get 
\[
\frac{\partial \phi(\widetilde{\boldsymbol{z}}^{(1)}|\boldsymbol{\theta}_{\cdot, 1},\boldsymbol{I}_{q})}{\partial \widetilde{\boldsymbol{z}}^{(1)}}=0 \;,\]
and equivalently
\[
\widetilde{\boldsymbol{z}}^{(1)}=\boldsymbol{\theta}_{\cdot, 1} \;,  
\quad \mbox{the only mode of a Gaussian density}.
\]
For any modes of $g(\widetilde{\boldsymbol{z}})$, the first part
$\widetilde{\boldsymbol{z}}^{(1)}$ all equal to
$\boldsymbol{\theta}_{\cdot, 1}$, that is, the projections of the
modes onto the null subspace $\boldsymbol{\nu}^{\perp}$ coincide at
$\theta_{\cdot,1}$. Hence the modes and component means lie in the 
same constrained subspace $\boldsymbol{\nu}$. 


\section*{Appendix B}\label{appendix2}
We prove Theorem 3.2 here. 
Assume 
$\boldsymbol{\nu}=\{\boldsymbol{v}_{q+1}, ..., \boldsymbol{v}_p\}$ is the constrained subspace where $\boldsymbol{\mu}_{kr}$'s reside, 
and $\boldsymbol{\nu}^{\perp}=\{\boldsymbol{v}_1, ..., \boldsymbol{v}_q\}$ is the corresponding null subspace, as defined in 
Section 3. 
We use the Bayes classification rule to classify a sample $x$: 
$\widehat{y}=\argmax_kf(Y=k|\mathbf{X}=\mathbf{x})=\argmax_kf(\mathbf{X}=\mathbf{x},Y=k)$.
\begin{equation}
f(\boldsymbol{X}=\boldsymbol{x},Y=k)=a_kf_k(\boldsymbol{x})\propto a_k \sum_{r=1}^{R_k}\pi_{kr}\exp(-(\boldsymbol{x}-\boldsymbol{\mu}_{kr})^t\boldsymbol{\Sigma}^{-1}(\boldsymbol{x}-\boldsymbol{\mu}_{kr}))\;. 
\label{eq:genmodel}
\end{equation}
Let $\boldsymbol{V}=\begin{pmatrix} \boldsymbol{v}_1^{t} \\ \vdots \\ \boldsymbol{v}_p^{t} \end{pmatrix}$. Matrix 
$\boldsymbol{V}$ is orthonormal because $\boldsymbol{v}_j$'s are orthonormal by construction. 
Consider the following cases of $\boldsymbol{\Sigma}$. 
\subsection*{B.1 $\boldsymbol{\Sigma}$ is an identity matrix}\label{append:b1}
From Eq. (\ref{eq:genmodel}), we have 
\begin{eqnarray}
&&\sum_{r=1}^{R_k}\pi_{kr}\exp(-(\boldsymbol{x}-\boldsymbol{\mu}_{kr})^t\boldsymbol{\Sigma}^{-1}(\boldsymbol{x}-\boldsymbol{\mu}_{kr})) \nonumber \\
&=& \sum_{r=1}^{R_k}\pi_{kr}\exp(-(\boldsymbol{x}-\boldsymbol{\mu}_{kr})^t(\boldsymbol{V}^{t}\boldsymbol{V})(\boldsymbol{x}-\boldsymbol{\mu}_{kr})) \nonumber \\
&=& \sum_{r=1}^{R_k}\pi_{kr}\exp(-(\boldsymbol{V}\boldsymbol{x}-\boldsymbol{V}\boldsymbol{\mu}_{kr})^t(\boldsymbol{V}\boldsymbol{x}-\boldsymbol{V}\boldsymbol{\mu}_{kr})) \nonumber \\
&=& \sum_{r=1}^{R_k}\pi_{kr}\exp(-\sum_{j=1}^p(\breve{x}_j-\breve{\mu}_{kr,j})^2)\;,
\label{eq:b1}
\end{eqnarray}
where 
$\breve{x}_j=\boldsymbol{v}^{t}_j\cdot \boldsymbol{x}$, $\breve{\mu}_{kr,j}=\boldsymbol{v}^{t}_j\cdot\boldsymbol{\mu}_{kr}$, $j=1,2,...,p$. 
Because $\breve{\mu}_{kr,j}=c_j$, identical across all $k$ and $r$ for $j=1,\cdots,q$,
the first $q$ terms in the sum of exponent in Eq. (\ref{eq:b1}) are all constants. We have 
\begin{eqnarray*}
&& \sum_{r=1}^{R_k}\pi_{kr}\exp(-\sum_{j=1}^p(\breve{x}_j-\breve{\mu}_{kr,j})^2) \\
&\propto& \sum_{r=1}^{R_k} \pi_{kr}\exp(-\sum_{j=q+1}^p(\breve{x}_j-\breve{\mu}_{kr,j})^2)\;.
\end{eqnarray*}
Therefore, 
\begin{eqnarray*}
f(\boldsymbol{X}=\boldsymbol{x},Y=k)\propto a_k \sum_{r=1}^{R_k} \pi_{kr}\exp(-\sum_{j=q+1}^p(\breve{x}_j-\breve{\mu}_{kr,j})^2)\;. 
\end{eqnarray*}
That is, to classify a sample $\boldsymbol{x}$, we only need the projection of $\boldsymbol{x}$ onto the constrained subspace
$\boldsymbol{\nu}^{\perp}=\{\boldsymbol{v}_1, ..., \boldsymbol{v}_q\}$. 

\subsection*{B.2 $\boldsymbol{\Sigma}$ is a non-identity matrix}
We can perform 
a linear transform (a ``whitening'' operation) on $\boldsymbol{X}$ so that the
transformed data have an identity covariance matrix $\boldsymbol{I}$. 
Find the matrix square root of $\boldsymbol{\Sigma}$, that is,
$\boldsymbol{\Sigma}=(\boldsymbol{\Sigma}^{\frac{1}{2}})^t
\boldsymbol{\Sigma}^{\frac{1}{2}}$.  If the eigen decomposition of
$\boldsymbol{\Sigma}$ is
$\boldsymbol{\Sigma}=V_{\boldsymbol{\Sigma}}D_{\boldsymbol{\Sigma}}
V_{\boldsymbol{\Sigma}}^{t}$, then 
$\boldsymbol{\Sigma}^{\frac{1}{2}}=D_{\boldsymbol{\Sigma}}^{\frac{1}{2}}V_{\boldsymbol{\Sigma}}^{t}$. 
Let
$\boldsymbol{Z}=(\boldsymbol{\Sigma}^{-\frac{1}{2}})^t\boldsymbol{X}$. The distribution of $\boldsymbol{Z}$ is
\begin{eqnarray*}
g(\boldsymbol{Z}=\boldsymbol{z}, Y=k)=a_k\sum_{r=1}^{R_k}\pi_{kr} \phi(\boldsymbol{z}|\tilde{\boldsymbol{\mu}}_{kr}, \boldsymbol{I})\;,
\end{eqnarray*}
where $\tilde{\boldsymbol{\mu}}_{kr}=(\boldsymbol{\Sigma}^{-\frac{1}{2}})^t\boldsymbol{\mu}_{kr}$. 
According to our assumption, $\boldsymbol{v}_j^t\cdot\boldsymbol{\mu}_{kr}=c_j$, i.e., identical
across all $k$ and $r$ for $j=1, ...,q$. Plugging into $\boldsymbol{\mu}_{kr}=(\boldsymbol{\Sigma}^{\frac{1}{2}})^t\tilde{\boldsymbol{\mu}}_{kr}$, 
we get $(\boldsymbol{\Sigma}^{\frac{1}{2}}\boldsymbol{v}_j)^t\cdot\tilde{\boldsymbol{\mu}}_{kr}=c_j$, $j=1,...,q$.  
This means for the transformed data, the component means $\tilde{\boldsymbol{\mu}}_{kr}$'s have a null space 
spanned by $\{\boldsymbol{\Sigma}^{\frac{1}{2}}\boldsymbol{v}_j|j=1, ..., q\}$. 
Correspondingly, the constrained subspace is spanned by
$\{(\boldsymbol{\Sigma}^{-\frac{1}{2}})^t \boldsymbol{v}_j|j=q+1,...,p\}$. 
It is easy to verify that the new null space and constrained subspace are orthogonal, since
$(\boldsymbol{\Sigma}^{\frac{1}{2}}\boldsymbol{v}_j)^{t} \cdot (\boldsymbol{\Sigma}^{-\frac{1}{2}})^{t}\boldsymbol{v}_{j'} = {\boldsymbol{v}_j}^t \cdot \boldsymbol{v}_{j'} = 0$, $j=1,...q$ and $j'=q+1,...,p$.
The spanning vectors for the constrained subspace, 
$(\boldsymbol{\Sigma}^{-\frac{1}{2}})^t \boldsymbol{v}_j$, $j=q+1,...,p$, are not orthonormal in general, but there exists
an orthonormal basis that spans the same subspace. 
With a slight abuse of notation, we use $\{(\boldsymbol{\Sigma}^{-\frac{1}{2}})^t\boldsymbol{v}_j|j=q+1,...,p\}$ to denote a $p\times (p-q)$ matrix containing the column vector $(\boldsymbol{\Sigma}^{-\frac{1}{2}})^t \boldsymbol{v}_j$. 
For any matrix $A$ of dimension $p\times d$, $d<p$, let the notation $orth(A)$ denote a $p\times d$ matrix 
whose column vectors are orthonormal and span the same subspace as the column vectors of $A$. 
According to \ref{append:b1}, for the transformed data $\boldsymbol{Z}$, we only need the projection of 
$\boldsymbol{Z}$ onto a subspace spanned by the column vectors of 
$orth(\{(\boldsymbol{\Sigma}^{-\frac{1}{2}})^t \boldsymbol{v}_j|j=q+1,...,p\})$ to compute the class posterior. 
Note that $\boldsymbol{Z}=(\boldsymbol{\Sigma}^{-\frac{1}{2}})^t\boldsymbol{X}$. So the subspace that matters for classification 
for the original data $\boldsymbol{X}$ is spanned by the column vectors of $(\boldsymbol{\Sigma}^{-\frac{1}{2}})\times orth(\{(\boldsymbol{\Sigma}^{-\frac{1}{2}})^t \boldsymbol{v}_j|j=q+1,...,p\})$.
Again, these column vectors are not 
orthonormal in general, but there exists
an orthonormal basis that spans the same subspace. This orthonormal basis is hence spanned by the column vectors of 
$orth((\boldsymbol{\Sigma}^{-\frac{1}{2}})\times orth(\{(\boldsymbol{\Sigma}^{-\frac{1}{2}})^t \boldsymbol{v}_j|j=q+1,...,p\}))$. 
Since 
$orth((\boldsymbol{\Sigma}^{-\frac{1}{2}})\times orth(\{(\boldsymbol{\Sigma}^{-\frac{1}{2}})^t \boldsymbol{v}_j|j=q+1,...,p\}))=orth(\{\boldsymbol{\Sigma}^{-1}\boldsymbol{v}_j|j=q+1,...,p\})$,\footnote{
Let matrix $A$ be a $p\times p$ square matrix and B be a $p\times d$ matrix, $d<p$, it
can be proved that
$orth(A\times orth(B))=orth(A\times B)$.} the subspace that matters for classification is thus spanned by the column vectors of 
$orth(\{\boldsymbol{\Sigma}^{-1}\boldsymbol{v}_j|j=q+1,...,p\})$. 

In summary, only the linear projection of the data onto a subspace with the same dimension as $\boldsymbol{\nu}$ matters for classification. 

\section*{Appendix C: Derivation of $\mu_{kr}$ in GEM}\label{append:c}
We derive the optimal $\boldsymbol{\mu}_{kr}$'s under constraint
(4) for a given $\boldsymbol{\Sigma}$. Note that the term in Eq. (6) 
that involves $\boldsymbol{\mu}_{kr}$'s is:
\begin{eqnarray}
-\frac{1}{2}\sum_{k=1}^K\sum_{r=1}^{R_k}\sum_{i=1}^{n_k}q_{i,kr}(\boldsymbol{x}_i-\boldsymbol{\mu}_{kr})^t\boldsymbol{\Sigma}^{-1}(\boldsymbol{x}_i-\boldsymbol{\mu}_{kr})\;.
\label{eq:mu}
\end{eqnarray}
Denote $\sum_{i=1}^{n_k}q_{i,kr}$ by $l_{kr}$. 
Let $\bar{\boldsymbol{x}}_{kr}=\sum_{i=1}^{n_k}q_{i,kr}\boldsymbol{x}_i/l_{kr}$, i.e., 
the weighted sample mean of the component $r$ in class $k$. 
To
maximize Eq. (\ref{eq:mu}) is equivalent to minimizing the following term~\citep{Anderson:2000}:
\begin{eqnarray}
\sum_{k=1}^K\sum_{r=1}^{R_k} l_{kr} (\bar{\boldsymbol{x}}_{kr}-\boldsymbol{\mu}_{kr})^t\boldsymbol{\Sigma}^{-1}(\bar{\boldsymbol{x}}_{kr}-\boldsymbol{\mu}_{kr}) \; .
\label{eq:newobj}
\end{eqnarray}
To solve the above optimization problem under constraint (\ref{eq:c}),
we need to find a linear transform such that in the transformed space,
the constraint is imposed on individual coordinates (rather
than linear combinations of them), and the objective function is a
weighted sum of squared Euclidean distances between the transformed
$\bar{\boldsymbol{x}}_{kr}$ and $\boldsymbol{\mu}_{kr}$.  Once this is
achieved, the optimal solution will simply be given by setting those unconstrained
coordinates within each component by the component-wise
sample mean, and the constrained coordinates by the
component-pooled sample mean. We will discuss the detailed solution in
the following.

Find the matrix square root of $\boldsymbol{\Sigma}$, that is,
$\boldsymbol{\Sigma}=(\boldsymbol{\Sigma}^{\frac{1}{2}})^t
\boldsymbol{\Sigma}^{\frac{1}{2}}$.  If the eigen decomposition of
$\boldsymbol{\Sigma}$ is
$\boldsymbol{\Sigma}=V_{\boldsymbol{\Sigma}}D_{\boldsymbol{\Sigma}}
V_{\boldsymbol{\Sigma}}^{t}$, then,
$\boldsymbol{\Sigma}^{\frac{1}{2}}=D_{\boldsymbol{\Sigma}}^{\frac{1}{2}}V_{\boldsymbol{\Sigma}}^{t}$.
Now perform the following change of variables: 
\begin{eqnarray}
&&\sum_{k=1}^K\sum_{r=1}^{R_k} l_{kr} (\bar{\boldsymbol{x}}_{kr}-\boldsymbol{\mu}_{kr})^t\boldsymbol{\Sigma}^{-1}(\bar{\boldsymbol{x}}_{kr}-\boldsymbol{\mu}_{kr}) \nonumber \\
&=& 
\sum_{k=1}^K\sum_{r=1}^{R_k} l_{kr} \left[
\left(\boldsymbol{\Sigma}^{-\frac{1}{2}}\right )^{t}(\bar{\boldsymbol{x}}_{kr}-\boldsymbol{\mu}_{kr})\right ]^t
\left[
\left(\boldsymbol{\Sigma}^{-\frac{1}{2}}\right )^{t}(\bar{\boldsymbol{x}}_{kr}-\boldsymbol{\mu}_{kr})\right ]  \nonumber \\
&=& \sum_{k=1}^K\sum_{r=1}^{R_k} l_{kr} (\tilde{\boldsymbol{x}}_{kr}-\tilde{\boldsymbol{\mu}}_{kr})^t 
(\tilde{\boldsymbol{x}}_{kr}-\tilde{\boldsymbol{\mu}}_{kr})\;,
\label{eq:obj2}
\end{eqnarray}
where $\tilde{\boldsymbol{\mu}}_{kr}=\left(\boldsymbol{\Sigma}^{-\frac{1}{2}}\right )^{t}\cdot\boldsymbol{\mu}_{kr}$, and
$\tilde{\boldsymbol{x}}_{kr}=\left(\boldsymbol{\Sigma}^{-\frac{1}{2}}\right )^{t}\cdot \bar{\boldsymbol{x}}_{kr}$.
Correspondingly, the constraint in (\ref{eq:c}) becomes
\begin{eqnarray}
\left(\boldsymbol{\Sigma}^{\frac{1}{2}} \boldsymbol{v}_j \right )^{t}\tilde{\boldsymbol{\mu}}_{kr}=\mbox{constant over
$r$ and $k$}\; , \quad j=1,...,q\;.
\label{eq:c2}
\end{eqnarray}
Let $\boldsymbol{b}_j=\boldsymbol{\Sigma}^{\frac{1}{2}} \boldsymbol{v}_j$ and $\boldsymbol{B}=(\boldsymbol{b}_1, \boldsymbol{b}_2, ...,
\boldsymbol{b}_q)$.  Note that the rank of $\boldsymbol{V} = (\boldsymbol{v}_1, ..., \boldsymbol{v}_q)$ is $q$.  Since
$\boldsymbol{\Sigma}^{\frac{1}{2}}$ is of full rank,
$\boldsymbol{B}=\boldsymbol{\Sigma}^{\frac{1}{2}}\boldsymbol{V}$ also has rank $q$.  The constraint in
(\ref{eq:c2}) becomes
\begin{eqnarray}
\boldsymbol{B}^t \tilde{\boldsymbol{\mu}}_{kr}=\boldsymbol{B}^t \tilde{\boldsymbol{\mu}}_{k'r'}\; ,\mbox{for any $r$, $r'=1, ..., R_k$, and any $k$, $k'=1,...,K$}\;.
\label{eq:c3}
\end{eqnarray}
Now perform a singular value decomposition (SVD) on $\boldsymbol{B}$, i.e., 
$\boldsymbol{B}=\boldsymbol{U_{B}} \boldsymbol{D_{B}} \boldsymbol{V_{B}}^{t}$,
where $\boldsymbol{V_{B}}$ is a $q\times q$ orthonormal matrix, 
$\boldsymbol{D_B}$ is a $q\times q$ diagonal matrix, which is non-singular
since the rank of $\boldsymbol{B}$ is $q$, and $U_B$ is a $p\times q$ orthonormal
matrix. Substituting the SVD of $\boldsymbol{B}$ in (\ref{eq:c3}), we get
\[
\boldsymbol{V_B} \boldsymbol{D_B} \boldsymbol{U_B}^{t}\tilde{\boldsymbol{\mu}}_{kr}=\boldsymbol{V_B} \boldsymbol{D_B} \boldsymbol{U_B}^{t}\tilde{\boldsymbol{\mu}}_{k'r'},\quad
\mbox{for any $r$, $r'=1, ..., R_k$, and any $k$, $k'=1,...,K$}\;,
\]
which is equivalent to 
\begin{eqnarray}
\boldsymbol{U_B}^{t}\tilde{\boldsymbol{\mu}}_{kr}=\boldsymbol{U_B}^{t}\tilde{\boldsymbol{\mu}}_{k'r'},\quad
\mbox{for any $r$, $r'=1, ..., R_k$, and any $k$, $k'=1,...,K$,}
\label{eq:c4}
\end{eqnarray}
because $\boldsymbol{V_B}$ and $\boldsymbol{D_B}$ have full rank.  We can augment $\boldsymbol{U_B}$ to a $p\times
p$ orthonormal matrix, $\hat{\boldsymbol{U}}=(\boldsymbol{u}_1, ..., \boldsymbol{u}_q, \boldsymbol{u}_{q+1}, ..., \boldsymbol{u}_{p})$, 
where $\boldsymbol{u}_{q+1}$, ..., $\boldsymbol{u}_p$ are augmented orthonormal vectors.  Since
$\hat{\boldsymbol{U}}$ is orthonormal, the objective function in Eq. (\ref{eq:obj2}) can be
written as
\begin{eqnarray}
\sum_{k=1}^K\sum_{r=1}^{R_k}l_{kr} [\hat{\boldsymbol{U}}^{t}(\tilde{\boldsymbol{x}}_{kr}-\tilde{\boldsymbol{\mu}}_j)]^t\cdot 
[\hat{\boldsymbol{U}}^{t}(\tilde{\boldsymbol{x}}_{kr}-\tilde{\boldsymbol{\mu}}_{kr})]
=\sum_{k=1}^K\sum_{r=1}^{R_k}l_{kr} (\breve{\boldsymbol{x}}_{kr}-\breve{\boldsymbol{\mu}}_{kr})^t (\breve{\boldsymbol{x}}_{kr}-\breve{\boldsymbol{\mu}}_{kr}) \; ,
\label{eq:obj3}
\end{eqnarray}
where $\breve{\boldsymbol{x}}_{kr}=\hat{\boldsymbol{U}}^t\tilde{\boldsymbol{x}}_{kr}$ and 
$\breve{\boldsymbol{\mu}}_{kr}=\hat{\boldsymbol{U}}^t\tilde{\boldsymbol{\mu}}_{kr}$. 
If we denote $\breve{\boldsymbol{\mu}}_{kr}=(\breve{\mu}_{kr,1}, \breve{\mu}_{kr,2}, ...,
\breve{\mu}_{kr,p})^t$, then the constraint in (\ref{eq:c4}) simply becomes
\begin{eqnarray*}
\breve{\mu}_{kr, j}=\breve{\mu}_{k'r', j}\;, \;
\mbox{for any $r$, $r'=1, ..., R_k$, and any $k$, $k'=1,...,K$, $j=1,...,q$}\;.
\end{eqnarray*}
That is, the first $q$ coordinates of $\breve{\boldsymbol{\mu}}$ have to be common over 
all the $k$ and $r$.  
The objective function (\ref{eq:obj3}) can be separated coordinate
wise:
\begin{eqnarray*}
\sum_{k=1}^{K}\sum_{r=1}^{R_k}l_{kr} (\breve{\boldsymbol{x}}_j-\breve{\boldsymbol{\mu}}_{kr})^t (\breve{\boldsymbol{x}}_{kr}-\breve{\boldsymbol{\mu}}_{kr})
=\sum_{j=1}^{p}\sum_{k=1}^{K}\sum_{r=1}^{R_k}l_{kr} (\breve{x}_{kr,j}-\breve{\mu}_{kr,j})^2 \;.
\label{eq:obj4}
\end{eqnarray*}
For the first $q$ coordinates, the optimal $\breve{\mu}_{kr,j}$,
$j=1,...,q$,  is
solved by
\begin{eqnarray}
\breve{\mu}_{kr,j}^{*}=\frac{\sum_{k'=1}^K\sum_{r'=1}^{R_{k'}}l_{k'r'} \breve{x}_{k'r',j}}
{\sum_{k'=1}^K\sum_{r'=1}^{R_{k'}}l_{k'r'}}=\frac{\sum_{k'=1}^K\sum_{r'=1}^{R_{k'}}l_{k'r'} \breve{x}_{k'r',j}}{n}
\; , \quad \mbox{identical over $r$ and $k$}\;. \nonumber
\end{eqnarray}
For the remaining coordinates, $\breve{\mu}_{kr,j}$, $j=q+1, ..., p$:
\begin{eqnarray}
\breve{\mu}_{kr,j}^{*}=\breve{x}_{kr,j} \;. \nonumber
\end{eqnarray}
After $\breve{\boldsymbol{\mu}}_{kr}^{*}$ is calculated, we finally get $\boldsymbol{\mu}_{kr}$'s under the constraint(\ref{eq:c}):
\begin{eqnarray}
\boldsymbol{\mu}_{kr}=(\boldsymbol{\Sigma}^{\frac{1}{2}})^t
\hat{\boldsymbol{U}} \breve{\boldsymbol{\mu}}_{kr}^{*} \;. \nonumber
\end{eqnarray}


\section*{Appendix D: Reduced Rank Mixture Discriminant Analysis}\label{append:d}
The rank restriction can be incorporated into the mixture
discriminant analysis (MDA). It is known that the rank-$L$ LDA fit is
equivalent to
a Gaussian maximum likelihood
solution, where the means of Gaussians lie in a
$L$-dimension subspace~\citep{Hastie:1996}. Similarly, in MDA, the log-likelihood
can be maximized with the restriction that all the $R=\sum_{k=1}^{K}R_k$
centroids are confined to a rank-$L$ subspace, i.e., rank $\{\mu_{kr}\}=L$.

The EM algorithm is used to estimate the parameters of the reduced rank MDA, and the M-step 
is a weighted version of LDA, with $R$
``classes''. The component posterior probabilities 
$q_{i,kr}$'s in
the E-step are calculated in the same way as in
Eq. (\ref{em:estep}), which are conditional on the current (reduced
rank) version of component means and common covariance matrix. In the
M-step, $\pi_{kr}$'s are still maximized using
Eq. (\ref{eq:prior}). The maximizations of $\boldsymbol{\mu}_{kr}$ and
$\boldsymbol{\Sigma}$ can be viewed as weighted mean and pooled
covariance maximum likelihood estimates in a weighted and augmented
$R$-class problem. Specifically, we augment the data by replicating
the $n_k$ observations in class $k$ $R_k$ times, with the $l$th such
replication having the observation weight $q_{i,kl}$.  This is done for
each of the $K$ classes, resulting in an augmented and weighted\textbf{}
training set of $\sum_{k=1}^Kn_kR_k$ observations. Note that the sum
of all the weights is $n$. We now impose the rank restriction.  For
all the sample points $\boldsymbol{x}_i$'s within class $k$, the
weighted component mean is
\begin{eqnarray*}
\boldsymbol{\mu}_{kr} = \frac{\sum_{i=1}^{n_k}q_{i,kr}\boldsymbol{x}_i}{\sum_{i=1}^{n_k}q_{i,kr}}\;. 
\label{eq:mukr}
\end{eqnarray*}
Let $q'_{kr}=\sum_{i=1}^{n_k}q_{i,kr}$. The overall mean is 
\begin{eqnarray*}
\boldsymbol{\mu} = \frac{\sum_{k=1}^{K}\sum_{r=1}^{R_k}q'_{kr}\boldsymbol{\mu}_{kr}}{\sum_{k=1}^{K}\sum_{r=1}^{R_k}q'_{kr}} \;.
\label{eq:allmu}
\end{eqnarray*}
The pooled within-class covariance matrix is 
\begin{eqnarray*}
W = \frac{\sum_{k=1}^{K}\sum_{r=1}^{R_k}\sum_{i=1}^{n_k}q_{i,kr}(\boldsymbol{x}_i-\boldsymbol{\mu}_{kr})^t(\boldsymbol{x}_i-\boldsymbol{\mu}_{kr})}
{\sum_{k=1}^{K}\sum_{r=1}^{R_k}q'_{kr}} \;.
\label{eq:withcov}
\end{eqnarray*}
The between-class covariance matrix is
\begin{eqnarray*}
B = \frac{\sum_{k=1}^{K}\sum_{r=1}^{R_k}q'_{kr}(\boldsymbol{\mu}_{kr}-\boldsymbol{\mu})^t(\boldsymbol{\mu}_{kr}-\boldsymbol{\mu})}{\sum_{k=1}^{K}\sum_{r=1}^{R_k}q'_{kr}}\;.
\label{eq:betweencov}
\end{eqnarray*}
Define $B^*=(W^{-\frac{1}{2}})^TBW^{-\frac{1}{2}}$. Now perform an eigen-decomposition on $B^*$, i.e., $B^*=V^*D_{B}{V^*}^T$, where 
$V^*=(v_1^{*},v_2^{*},...,v_p^{*})$. Let $V$ be a matrix consisting of the leading $L$ columns of $W^{-\frac{1}{2}}V^*$. Considering 
maximizing the Gaussian log-likelihood subject to the constraints rank $\{\boldsymbol{\mu}_{kr}\}=L$, the solutions are
\begin{eqnarray}
\hat{\boldsymbol{\mu}}_{kr}=WVV^{T}(\boldsymbol{\mu}_{kr}-\boldsymbol{\mu})+\boldsymbol{\mu} \;,
\label{reducedmean}
\end{eqnarray}
\begin{eqnarray}
\hat{\boldsymbol{\Sigma}}=W+\frac{\sum_{k=1}^{K}\sum_{r=1}^{R_k}q'_{kr}(\boldsymbol{\mu}_{kr}-\hat{\boldsymbol{\mu}}_{kr})^t(\boldsymbol{\mu}_{kr}-\hat{\boldsymbol{\mu}}_{kr})}{\sum_{k=1}^{K}\sum_{r=1}^{R_k}q'_{kr}}\;.
\label{reducedsigma}
\end{eqnarray}

As a summary, in the M-step of reduced rank MDA, the parameters,
$\pi_{kr}$, $\boldsymbol{\mu}_{kr}$ and $\boldsymbol{\Sigma}$, are
maximized by Eqs. (\ref{eq:prior}), (\ref{reducedmean}), and
(\ref{reducedsigma}), respectively.

Note that the discriminant subspace is spanned by the column vectors of $V=W^{-\frac{1}{2}}V^*$, with the $l$th discriminant variable as $W^{-\frac{1}{2}}v_l^*$. In general, $W^{-\frac{1}{2}}v_l^*$'s are not orthogonal, but we can find an orthonormal basis that spans the 
same subspace. 

\end{document}